\documentclass[aps,
prx,
twocolumn,
superscriptaddress,
longbibliography,
preprintnumbers,
10pt,
floatfix]{revtex4-2}
\usepackage{amsmath}
\usepackage{amssymb}
\usepackage{color}
\usepackage[table]{xcolor}
\usepackage{textcomp}
\usepackage{xspace,slashed}
\usepackage[utf8]{inputenc}
\usepackage{multirow}
\usepackage{nccmath}
\usepackage{slashed}
\usepackage{orcidlink}
\usepackage{booktabs} 
\usepackage{longtable}
\usepackage{listings} %
\usepackage{hyperref}
\makeatletter
\AtBeginDocument{\let\LS@rot\@undefined}

\renewcommand{\appendix}{%
  \par
  \setcounter{section}{0}%
  \setcounter{subsection}{0}%
  \gdef\@chapapp{\appendixname}%
  \gdef\thesection{\Alph{section}}%
}
\makeatother

\newtheorem{proof}{Proof}

\usepackage{amsfonts}
\usepackage{graphicx}
\usepackage{dcolumn}
\usepackage{bm}
\usepackage{url}

\newcommand{\beginsupplement}{%
    \setcounter{table}{0}%
    \renewcommand{\thetable}{S\arabic{table}}%
    \renewcommand{\theHtable}{S\arabic{table}}%
    \setcounter{figure}{0}%
    \renewcommand{\thefigure}{S\arabic{figure}}%
    \renewcommand{\theHfigure}{S\arabic{figure}}%
    \setcounter{equation}{0}%
    \renewcommand{\theequation}{S\arabic{equation}}%
    \renewcommand{\theHequation}{S\arabic{equation}}%
    \setcounter{section}{0}%
    \renewcommand{\thesection}{S\arabic{section}}%
    \renewcommand{\theHsection}{S\arabic{section}}%
    \renewcommand{\thesubsection}{\thesection.\Alph{subsection}}%
    \setcounter{page}{1}%
    \renewcommand{\thepage}{S\arabic{page}}%
}

\begin{document}

\title{Detailed Balance in Large-Language-Model-Driven Agents}

\author{Zhuo-Yang Song~\orcidlink{0009-0001-9727-7908}}
\email{zhuoyangsong@stu.pku.edu.cn}
\affiliation{School of Physics, Peking University, Beijing 100871, China}

\author{Qing-Hong Cao}
\email{qinghongcao@pku.edu.cn}
\affiliation{School of Physics, Peking University, Beijing 100871, China}
\affiliation{School of Physics, Zhengzhou University, Zhengzhou 450001, China}
\affiliation{Center for High Energy Physics, Peking University, Beijing 100871, China}

\author{Ming-xing Luo}
\email{mingxingluo@csrc.ac.cn}
\affiliation{Beijing Computational Science Research Center, Beijing 100193, China}

\author{Hua Xing Zhu~\orcidlink{0000-0002-7129-6748}}
\email{zhuhx@pku.edu.cn}
\affiliation{School of Physics, Peking University, Beijing 100871, China}
\affiliation{Center for High Energy Physics, Peking University, Beijing 100871, China}

\begin{abstract}
Large language model (LLM)-driven agents are emerging as a powerful new paradigm for solving complex problems. Despite the empirical success of these practices, a theoretical framework to understand and unify their macroscopic dynamics remains lacking. This paper proposes a method based on the least action principle to estimate the underlying generative directionality of LLMs embedded within agents. By experimentally measuring the transition probabilities between LLM-generated states, we report evidence consistent with a tendency toward detailed balance in LLM-generated transitions, indicating that LLM generation may not be achieved by generally learning rule sets and strategies, but rather by implicitly learning a class of underlying potential functions that may transcend different LLM architectures and prompt templates. To our knowledge, this provides an effective macroscopic description of LLM generative dynamics that does not depend on specific model details. This work is an attempt to establish a macroscopic dynamics theory of complex AI systems, aiming to elevate the study of AI agents from a collection of engineering practices to a science built on effective measurements that are predictable and quantifiable.
\end{abstract}

\maketitle

\section{Introduction} Large language model (LLM)-driven agents are emerging as a powerful new paradigm for solving complex problems~\cite{NEURIPS2022_9d560961,yao2023reactsynergizingreasoningacting,yao2023treethoughtsdeliberateproblem,wang2023selfconsistencyimproveschainthought,wang2023planandsolvepromptingimprovingzeroshot,schick2023toolformerlanguagemodelsteach,wang2023voyageropenendedembodiedagent,chen2023programthoughtspromptingdisentangling,pmlr-v202-gao23f,liang2023codepolicieslanguagemodel,mankowitz2023faster,Park2023,Besta_Blach_Kubicek_Gerstenberger_Podstawski_Gianinazzi_Gajda_Lehmann_Niewiadomski_Nyczyk_Hoefler_2024,Mohammadi_2025,sapkota2025aiagentsvsagentic}, demonstrating potential in frontier areas such as scientific discovery by combining the generative capabilities of LLMs with external tools and memory systems~\cite{boiko2023autonomous,bran2023chemcrow,funsearch, alphaevolve,song2025iteratedagentsymbolicregression}. For instance, FunSearch and AlphaEvolve achieve iterative optimization of solutions by integrating LLMs into evolutionary algorithm frameworks~\cite{funsearch, alphaevolve}. However, the theoretical understanding and explanation of LLMs often remain at the level of token statistical properties and microscopic generative mechanisms~\cite{Bhattacharya2022,sunyuran2025phasetransitions,liu2025neuralthermodynamiclawslarge}, making it difficult to explain the macroscopic dynamics of LLMs as complex systems~\cite{Hoel2017WhenTM,NEURIPS2022_9d560961}. The behavior of these LLM-driven agents is often viewed as a direct product of their complex internal engineering (such as prompt templates, memory modules, tool calls), and their dynamic characteristics remain a black box~\cite{Hopfield1982,wei2022emergent,schaeffer2023are,Park2023,kim2025sciencescalingagentsystems}.

The dynamics of LLM generation are quite unique. Compared to traditional rule-based programs, LLM-based generation exhibits diverse and adaptive outputs~\cite{Hopfield1982,boiko2023autonomous,bran2023chemcrow}. At the same time, compared to naive random search, LLM generation shows stronger structure and goal-orientedness~\cite{hu2025conceptnetwork,funsearch,alphaevolve,song2025iteratedagentsymbolicregression}. Despite the complexity of this hybrid dynamics between random search and deterministic planning, we show that at the agent level (i.e., a coarse-grained description of LLM generative dynamics with standardized agent states as units), LLM generative dynamics exhibit detailed balance similar to equilibrium systems, thereby greatly simplifying the analysis and understanding of LLM generative dynamics~\cite{Markov2006ExtensionOT,MC1953,SpinGlassBeyond,schroeder2020introduction}.

To model the dynamic behavior of LLMs, we embed the generative process of LLM within a given agent framework, viewing it as a Markov transition process in its state space~\cite{Markov2006ExtensionOT,MC1953,Park2023,wang2023voyageropenendedembodiedagent,funsearch}. The states are defined by the complete information retained by the agent at each time step, which may include task objectives, historical summaries, code, file systems, application programming interface return values, etc., where the LLM-based generative process is treated as the transition kernel from the current state to a new state. We show that even if the states do not contain complete historical records (as in, e.g., FunSearch~\cite{funsearch}), the LLM-based state transitions can still exhibit a directionality toward specific states. We attribute this characteristic to the state transitions approximately satisfying the detailed balance condition, a nontrivial constraint that rules out probability-flux cycles; from it a task-specific potential function $V$ emerges that evaluates the intrinsic properties of any given state, such as ``how far the LLM perceives it to be from the goal.'' This global awareness enables LLMs to quickly converge to those optimal states, effectively avoiding repetitive cycles in the state space and achieving stronger generalization capabilities than merely learning strategy sets~\cite{sutton1998reinforcement,Tishby2015DeepLearning,Keskar2016OnLT}.

Based on this model, we propose a method to measure this underlying potential function based on a least action principle~\cite{Goldstein2002Classical,Friston2010,Tschantz2020scaling}. By experimentally measuring the transition probabilities between states, we report evidence consistent with a tendency toward detailed balance in LLM-generated transitions, indicating that LLM generation may not be achieved by generally learning rule sets and strategies, but rather by implicitly encoding an underlying potential function that transcends different LLM architectures and prompt templates. We propose detailed balance as a universal conjecture about LLM-driven generation; the potential $V$ is a task-specific effective description. This work is an attempt to establish a macroscopic dynamics theory of complex AI systems, aiming to elevate the study of AI agents from a collection of engineering practices to a predictable and quantifiable science built on effective measurements.

\section{Theory} 
\begin{figure}[tbp]
    \centering
    \includegraphics[width=0.9\linewidth]{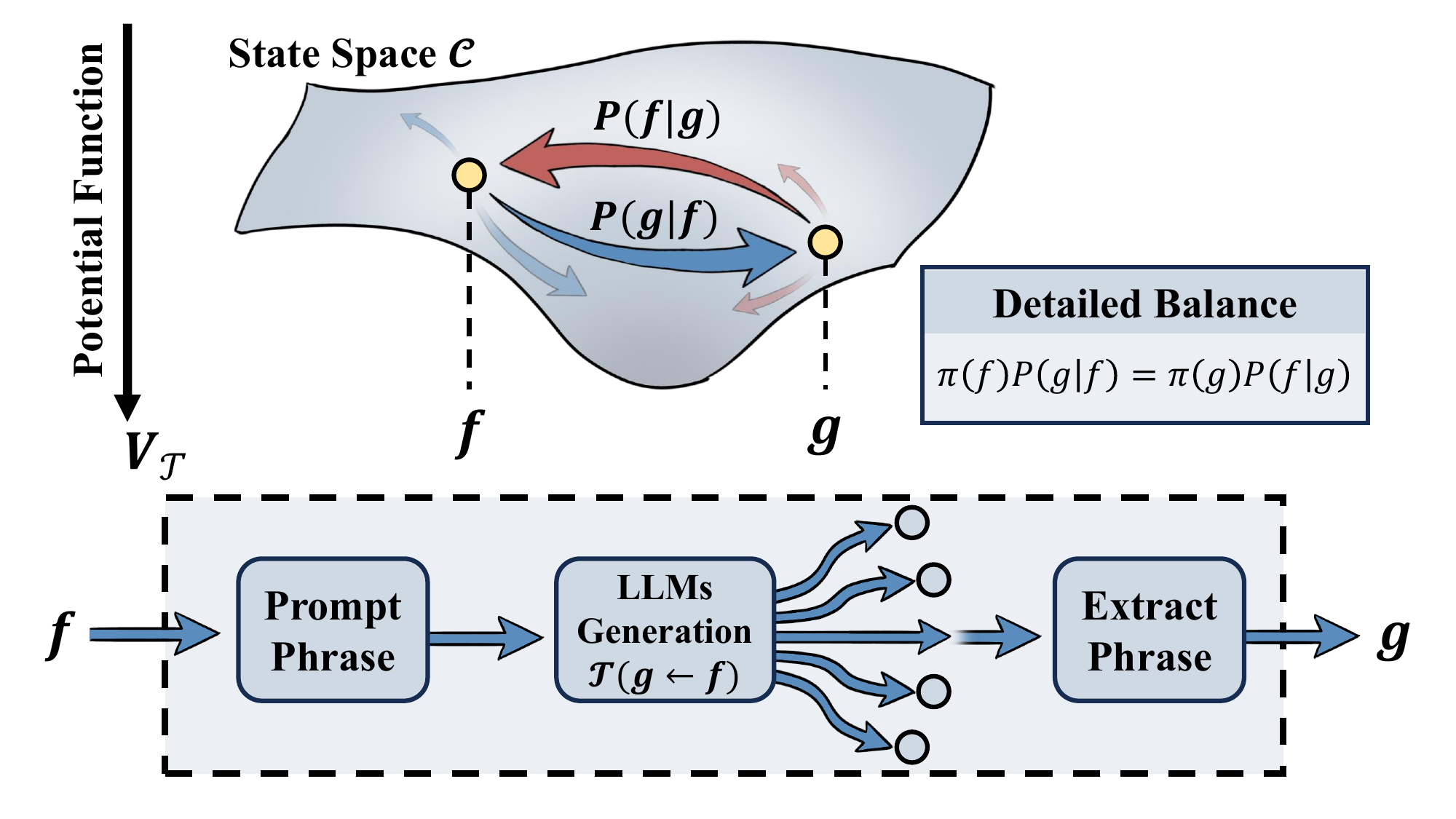}
    \caption{A schematic of a formalization framework for studying the directionality of LLM generation, illustrating the state space and possible transitions. LLMs are embedded within an agent, which transitions from state $f$ to state $g$ with probability $\mathcal{T}(g\gets f) = P(g|f)$. The underlying potential function $V_{\mathcal{T}}$ quantifies the agent's global ordering of each state, satisfying the detailed balance condition at equilibrium.}
    \label{fig:theory}
\end{figure}

To formulate the problem rigorously, we consider an agent whose core consists of one or more LLMs. The agent takes its current state $f$ as input. Through a series of deterministic steps, it organizes and evaluates this state to generate a relevant prompt. This prompt is then fed into one or more LLMs, whose structured output is parsed to get a new state $g$. This state is the minimal unit for studying LLM dynamics. This generative process can be viewed as a Markov transition process in the state space $\mathcal{C}$ with a transition kernel $P(g|f)$, retaining the diversity and adaptability of LLM generation. The states are defined by the complete information retained by the agent at each time step, which should include all the information required for the agent to carry out a continuous reasoning or analogical process~\cite{west2001introduction,yasunaga2024largelanguagemodelsanalogical,funsearch,song2025iteratedagentsymbolicregression}. In this paper, the agent contains only a single generation step of LLM, and we denote $\mathcal{T}(g\gets f) = P(g|f)$ as the probability of the agent transitioning from a template containing state $f$ to an output containing state $g$ through LLM generation. A schematic diagram is shown in Fig.~\ref{fig:theory}.

LLM-based agents are characterized by their state transitions not being entirely random but exhibiting a certain structured preference. Specifically, agents tend to transition from the current state $f$ to states $g$ that are ``better'' from the agent's perspective. To capture this phenomenon, we hypothesize the existence of an underlying potential function $V_{\mathcal{T}}:\mathcal{C}\to\mathbb{R}$, which assigns a scalar value to each state, reflecting its ``quality''. Since a specific potential function is often difficult to compute directly, we propose a method to effectively estimate the potential function.

Given a global potential function $V$, we define the violation of the agent's given transition $\mathcal{T}(g\gets f)$ to the potential function as $K\!\left(V(f)-V(g)\right)$, where $K(x)$ is a convex function that describes the extent to which the transition from state $f$ to state $g$ violates the ordering of the potential function $V$. To quantify the overall mismatch between the agent's behavior and the potential function, we weight by the transition kernel $\mathcal{T}(g\gets f)$ and define the action $\mathcal{S}$ as the global average violation:
\begin{align}
  \mathcal{S} = \int_{f\in \mathcal{C}}\int_{g\in \mathcal{C}} \mathcal{T}(g\gets f)\, K\!\left(V(f)-V(g)\right)\, Df\, Dg,
  \label{eq:def:S}
\end{align}
where $Df,Dg$ are measures on the state space ($\int Df \equiv \sum_f$ for the discrete spaces in this paper). We choose $K(x)=\exp(-\beta x/2)$ as the convex function describing the violation of the given state transition from $f$ to $g$ in the ordering of the scalar function $V$. Here, $\beta$ is a scaling factor; for a single agent it can be absorbed into $V$, but when comparing across agents it acquires physical meaning through majority voting (see Sec.~S4 of the Supplemental Material~\cite{song2026supplement}). The action $\mathcal{S}$ or the distribution shape of $\beta V(f)$ can represent the agent's global cognition ability within this state space $\mathcal{C}$.
We propose that to quantify the behavior of LLMs using a potential function, one can seek such a potential function that minimizes the overall mismatch between the agent's transitions and the potential function~\cite{Friston2010,Tschantz2020scaling}. Therefore, the most suitable potential function $V_{\mathcal{T}}$ for describing an LLM-based agent $\mathcal{T}$ in a given state space is the one that minimizes the action $\mathcal{S}$~\cite{west2001introduction,optimize2002,Goldstein2002Classical,Jiang2011graph}.

This implies that the action satisfies the variational principle with respect to the potential function $V_{\mathcal{T}}$\footnote{We show in Sec.~S1 of the Supplemental Material~\cite{song2026supplement} that the variational principle is equivalent to the least action principle under the condition that $K(x)$ is a convex function.}~\cite{Goldstein2002Classical}:
\begin{align}
  \delta \mathcal{S} = 0.
  \label{eq:def:minimumS}
\end{align}

The variational condition is equivalent to $V_{\mathcal{T}}$ satisfying the following equilibrium condition:
\begin{align}
  &\int_{g \in \mathcal{C}} \mathcal{T}(g\gets f)\, K'\!\left(V_{\mathcal{T}}(f)-V_{\mathcal{T}}(g)\right)\, Dg\nonumber\\
  -& \int_{h \in \mathcal{C}} \mathcal{T}(f\gets h)\, K'\!\left(V_{\mathcal{T}}(h)-V_{\mathcal{T}}(f)\right)\, Dh = 0,
  \label{eq:minV}
\end{align}
holding for all $f \in \mathcal{C}$, where $K'(x)=\frac{dK}{dx}$.

Specifically, if for all transitions $\mathcal{T}(g\gets f)>0$, $V(f) \geq V(g)$ holds, it indicates that the agent's state transitions are completely ordered, and in this case, $V$ serves as a Lyapunov function~\cite{lyapunov1992general,strogatz2018nonlinear}.

It is worth noting that if $\mathcal{T}$ describes the transition of an equilibrium system, its state transitions satisfy the detailed balance condition, i.e., for all state pairs $(f,g)$, the following holds~\cite{MC1953,schroeder2020introduction}:
\begin{align}
    \pi(f) P(g|f) = \pi(g) P(f|g),
    \label{eq:detailed balance def}
\end{align}
where $\pi(f)$ denotes the equilibrium distribution of the system at state $f$, and $P(g|f)$ denotes the transition kernel. In this case, there exists a potential function $V$ that can explicitly express the detailed balance as
\begin{align}
  \log{\frac{\mathcal{T}(g\gets f)}{\mathcal{T}(f\gets g)}} = \beta V(f) - \beta V(g).
  \label{eq:detailed balance}
\end{align}
Unless otherwise specified, $\log$ denotes the natural logarithm (base $e$).
Substituting into Eq.~\eqref{eq:minV}, it can be verified that this potential function $V=V_{\mathcal{T}}$ satisfies the least action principle. This indicates that for equilibrium systems, if the detailed balance condition exists, the corresponding underlying potential function can be estimated through the least action principle. In general cases, the least action merely seeks the most ordered arrangement of the potential function, minimizing the violations of this arrangement by the agent's state transitions~\cite{Jiang2011graph}.

The Markov formulation is well motivated: LLM-agent systems routinely run far beyond the context window~\cite{funsearch,alphaevolve,song2025iteratedagentsymbolicregression,wang2023voyageropenendedembodiedagent,Park2023,meyerson2025solvingmillionstepllmtask}, necessitating a distill-and-generate protocol that makes the memoryless approximation a practical description of the dynamics.\footnote{With the Markov chain defined, time-reversal symmetry is well defined on its transition kernel as the equality of forward and reverse transition weights around every closed path, equivalent to detailed balance.}

The central point of this paper is that LLM-based agents often behave like an equilibrium system in their LLM-generated state space, which is coarse-grained compared to the complete generation sequence of LLMs~\cite{weinberg_1979,Hoel2017WhenTM}. The existence of this phenomenon suggests a universal macroscopic law in LLM generative dynamics that does not depend on specific model or task details. It indicates that despite being seemingly unrelated, there are underlying connections between different LLM generative processes, allowing us to describe the global ordering in LLM generation through the potential function $V_{\mathcal{T}}$, thereby providing explanations for the internal dynamics of LLMs.

\section{Experiments} We conducted experiments on three different models, including GPT-5 Nano, Claude Sonnet 4.5, and Gemini 2.5 Flash~\cite{gpt5systemcard,claude4systemcard,geminimodelcard}. Each model was prompted to generate a new word based on a given prompt word such that the sum of the letter indices of the new word equals 100. For example, given the prompt ``WIZARDS(23+9+26+1+18+4+19=100)'', the model needs to generate a new word whose letter indices also sum to 100, such as ``BUZZY(2+21+26+26+25=100)''. The transition kernel between two prompt words can be estimated through sampling as:
\begin{align}
    \mathcal{T}(g\gets f) \approx \frac{N(g\gets f)}{N_0(f)},
\end{align}
where $N(g\gets f)$ denotes the number of times the model generated the word $g$ from the prompt word $f$. $N_0(f)$ represents the number of sampling attempts starting from the prompt word $f$. Each model performed 20,000 generations. More details of the experiments are provided in Sec.~S2 of the Supplemental Material~\cite{song2026supplement}.

The three models exhibited two different behaviors, demonstrating directionality and certain diversity in actual LLM generative dynamics. Claude Sonnet 4.5 and Gemini 2.5 Flash demonstrated rapid convergence, with generated words quickly concentrating on a few high-frequency words. For instance, in 20,000 generations, Claude Sonnet 4.5 generated only 5 valid prompt words, while Gemini 2.5 Flash generated 13 valid prompt words. In contrast, GPT-5 Nano exhibited stronger exploration, producing as many as 645 different valid prompt words in 20,000 generations. This difference reflects the exploration-exploitation trade-off in LLM generative dynamics, making them suitable for task scenarios with varying demands for exploration and stability.

In Claude Sonnet 4.5 and Gemini 2.5 Flash, the solution to the variational condition Eq.~\eqref{eq:def:minimumS} can be calculated analytically. In this case, we can plot the transition process of Claude Sonnet 4.5 ordered by the potential function as shown in Fig.~\ref{fig:transition_matrix}, with specific calculations provided in Sec.~S3 of the Supplemental Material~\cite{song2026supplement}.

\begin{figure}
    \centering
    \includegraphics[width=0.8\linewidth]{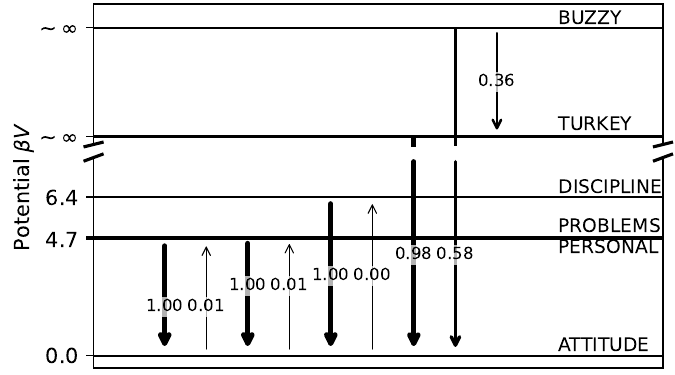}
    \caption{In the conditioned word generation task, the Claude Sonnet 4.5 model exhibits directionality. Transition process of the Claude Sonnet 4.5 model in the prompt word state space, ordered by the potential function $V_{\mathcal{T}}$. Transitions tend to move toward states with lower potentials. States with $\beta V(f)\gg \log(20000)\sim 10$ are those where the equilibrium condition cannot be strictly satisfied.}
    \label{fig:transition_matrix}
\end{figure}

Since Claude Sonnet 4.5 and Gemini 2.5 Flash exhibited high convergence, the equilibrium condition Eq.~\eqref{eq:minV} is almost equivalent to the detailed balance condition Eq.~\eqref{eq:detailed balance}. It is worth noting that we observed Claude Sonnet 4.5 starting from the prompt word ``ATTITUDE'', which has the lowest potential function, began to attempt some invalid words, while Gemini 2.5 Flash oscillated between the two lowest potential functions ``ATTITUDE'' and ``DISCIPLINE'', losing exploration. This behavior is similar to the low-temperature trapping phenomenon in physical systems~\cite{schroeder2020introduction,J1997}, suggesting that controlling the potential function may provide a feasible path to avoid model convergence.

A key example of interest is the GPT-5 Nano model. GPT-5 Nano generated a large number of prompt words due to its strong exploration, allowing us to directly test the detailed balance condition within the state space. We note that according to detailed balance, the sum of potential changes along any closed path should be zero. Specifically, considering a closed path $f_1\to f_2\to \cdots \to f_n \to f_1$, according to the detailed balance condition, we have:
\begin{align}
  \sum_{i=1}^{n} \log{\frac{\mathcal{T}(f_{i+1}\gets f_i)}{\mathcal{T}(f_i\gets f_{i+1})}} = \sum_{i=1}^{n} \beta \left(V(f_i)-V(f_{i+1})\right) = 0,
  \label{eq:loop}
\end{align}

\begin{figure}[tbp]
    \centering
    \includegraphics[width=0.9\linewidth]{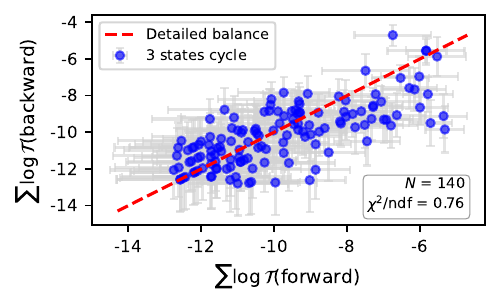}
    \caption{In the task of conditioned word generation without a reasoning chain, verification of detailed balance through closed paths in the state transition graph of the GPT-5 Nano model. Each point represents a triplet, with all different triplets found in the experimental data. The error bars directly arise from sampling errors. $\chi^2/\text{ndf}$ is the chi-square per degree of freedom. Note that both axes are in natural logarithm ($\ln$), not $\log_{10}$.}
    \label{fig:closed loop word}
\end{figure}

Figure~\ref{fig:closed loop word} counts all triplets in the experimental data, where transitions between each pair of the three data points were measured, totaling 140 different triplets. Each point represents a comparison of the sum of the logarithms of the forward and reverse transition kernels for a triplet, thereby testing the detailed balance condition. The measurement points cluster around the diagonal line, indicating that within the error range, the two sums are approximately equal, consistent with the detailed balance condition.

\begin{table}[tbp]
\caption{Examples of some states of the agent $\mathcal{T}_{\text{real}}$ and their potentials. The independent variable \texttt{log\_v\_k\_nu} in the fitting task is abbreviated as \texttt{x}.}
\begin{ruledtabular}
\begin{tabular}{lc}
 \multicolumn{1}{c}{states $f$} & \multicolumn{1}{c}{Potential}\\ \hline
\verb|param1 * tanh(param2 * x + param3) + param4| &  5.70\\
\verb|param1 - (param2 / (x + param3))| &  0.88\\
\verb|param1 * x / (1 + param2 * log(x + 1))| & -0.57\\
\verb|param1 * tanh(param2 * x) + param3|& -1.57\\
\verb|param2 + param1 * (1 - exp(-x))| & -3.30\\
\end{tabular}
\label{tab:states}
\end{ruledtabular}
\end{table}

To further test the universality of detailed balance in LLM generation, we now construct an agent $\mathcal{T}_{\text{real}}$ with a long reasoning chain, whose states are strings that can be parsed into specific expression trees. We recorded 50,228 state transitions executed by this agent, constructing a database containing 21,697 different transitions and 7,484 different states. Some example states are shown in Table~\ref{tab:states}. By analyzing these transitions, we can statistically test the detailed balance condition and estimate its underlying potential function $V_{\mathcal{T_{\text{real}}}}$ through the least action principle. This agent involves multiple different LLMs and prompt templates, so its potential function characteristics may reflect typical behaviors of LLM generation in practical applications. More experimental details are provided in Sec.~S2 of the Supplemental Material~\cite{song2026supplement}.

Similar to the measurement described above, we counted all triplets in the experimental data, where transitions between each pair of the three data points were measured, and for all different triplets, we observed consistency with the detailed balance condition within the measurement error range. The results are provided in Sec.~S2 of the Supplemental Material~\cite{song2026supplement}.

We now estimate the underlying potential function $V_{\mathcal{T_{\text{real}}}}$ through the least action principle. In a discrete state space, the integrals are replaced by sums over states, so the action (Eq.~\eqref{eq:def:S}) becomes, normalized by the number of states in the database:
\begin{align}
  \mathcal{S} = \frac{\sum_{g\gets f} K\!\left(V_{\mathcal{T_{\text{real}}}}(f) - V_{\mathcal{T_{\text{real}}}}(g)\right)}{\sum_{f} 1}.
  \label{eq:expS}
\end{align}
By numerically minimizing this action, we can estimate the potential function values $V_{\mathcal{T}_{\text{real}}}(f)$ for each state. According to this estimation, the minimum value of the action is much smaller than $K(0)$, indicating that the state transitions of the agent $\mathcal{T}_{\text{real}}$ indeed exhibit directionality. In Sec.~S4 of the Supplemental Material~\cite{song2026supplement}, we show that this optimized action value can be used to quantify the density distribution of states with respect to the underlying potential function, thereby providing direct guidance for designing more effective LLM generation strategies in practice.

By estimating this potential function, we can check whether the potential function is consistent with the detailed balance condition. Specifically, the detailed balance condition requires that for all state pairs $(f,g)$, Eq.~\eqref{eq:detailed balance} holds.

\begin{figure}[tbp]
    \centering
    \includegraphics[width=0.9\linewidth]{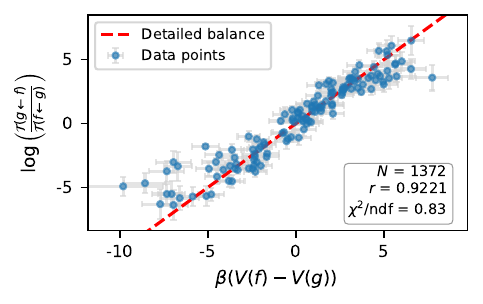}
    \caption{In the task of symbolic fitting with a long reasoning chain, test of the detailed balance condition for the agent $\mathcal{T}_{\text{real}}$. The error estimation includes root mean square statistical errors, and the error of the potential function is estimated through bootstrapping. $r$ is the Pearson correlation coefficient, and $\chi^2/\text{ndf}$ is the chi-squared per degree of freedom. Measurement includes state pairs with at least one measured transition between the two states. Points with $|V_{\mathcal{T}}(f)-V_{\mathcal{T}}(g)|>\log{50288}$ are excluded. One-tenth of the data points are displayed for clarity.}
    \label{fig:results}
\end{figure}

Figure~\ref{fig:results} shows a comparison of the left and right sides of Eq.~\eqref{eq:detailed balance}. We use the Pearson correlation coefficient $r$ between the two sides as the primary quantitative measure of how well detailed balance is satisfied; here $r=0.92$, indicating that the detailed balance condition holds to good approximation.

With the estimation of the potential function, we further discuss the specific meaning of the potential function in this problem to reveal the intrinsic cognitive characteristics in LLM generative dynamics. To this end, we use the action as an optimization objective and employ a workflow based on IdeaSearch~\cite{ideasearch2025} to find a potential function with an explicit functional form that maps the expression strings corresponding to state $f$ to scalar potential function values $V_{\mathcal{T}}(f)$. The best potential function found in 4000 rounds of search contains 49 parameters, capturing various features of state $f$ at the expression level, such as complexity, syntactic validity, and structural affinity with domain-specific patterns, without capturing string-level information. The magnitude of the corresponding parameter values directly reflects the importance that LLMs attach to these features during the generation process. The potential function and its specific analysis are provided in Sec.~S5 of the Supplemental Material~\cite{song2026supplement}.

\begin{figure}[tbp]
    \centering
    \includegraphics[width=0.9\linewidth]{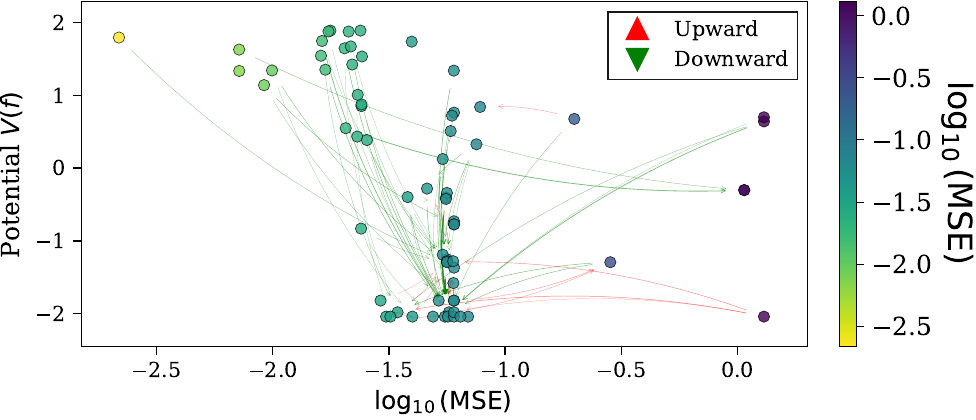}
    \caption{Illustration of the ability of the potential function discovered using IdeaSearch~\cite{ideasearch2025} to predict the directionality of state transitions. Each point represents a transition pair from state $f$ to state $g$. The figure shows a subgraph composed of 70 states selected from the database, displaying transitions with high transition kernel $\mathcal{T}_{\text{real}}(g\gets f)>0.05$. Red and green lines represent transitions with increasing and decreasing potential functions, respectively. The horizontal axis represents the mean square error of the expression corresponding to the state in the symbolic fitting task, while the vertical axis represents the potential function.}
    \label{fig:ideasearch}
\end{figure}

Figure~\ref{fig:ideasearch} shows the transition patterns between some states sorted by this potential function. In the database, there are a total of 9,769 transitions with high transition kernel $\mathcal{T}_{\text{real}}(g\gets f)>0.05$, and the corresponding potentials $V(f),V(g)$ are calculated using the potential function. Among them, 6,795 (69.56\%) exhibit a decrease in the potential function, 2,523 (25.83\%) exhibit an increase in the potential function, and 451 (4.62\%) exhibit no change in the potential function. This indicates that the potential function partially captures the intrinsic directionality in LLM generative dynamics. It is worth emphasizing that LLMs overall tend to choose states with relatively low potential function values as the next state, even though they may not necessarily perform better on actual data. In this way, the potential function can reveal the differences between the intrinsic cognition of LLMs and real data. Additional experiments on diverse tasks further support the generality of this conclusion (see Sec.~S2 of the Supplemental Material~\cite{song2026supplement}).

\section{Conclusion and Outlook} In this paper, we have proposed a framework based on the least action principle to describe and analyze the generative dynamics of LLM-based agents in their LLM-generated state space. Through experiments on multiple different models and tasks, we have found that the state transitions of these agents largely satisfy the detailed balance condition, indicating that their generative dynamics exhibit characteristics similar to equilibrium systems. We have further estimated the underlying potential function through the least action principle and revealed its important role in capturing the intrinsic directionality in LLM generative dynamics.

This paper provides a preliminary exploration of the possibility for discovering macroscopic laws in LLMs generative dynamics. Future work can further expand this framework and explore the application potential of more tools from equilibrium and near-equilibrium systems in understanding and optimizing LLM generation processes. For instance, studying the degree of deviation from equilibrium may help us understand a model's level of overfitting, as overfitted models may learn more localized strategy sets rather than global generative patterns governed by potential functions~\cite{Tishby2015DeepLearning,Keskar2016OnLT}. Additionally, optimization methods based on potential functions may also provide new ideas for improving the quality and diversity of LLM task-related generation, such as adjusting the action to different magnitudes based on varying safety and exploration requirements.

\begin{acknowledgments}
We would like to thank Z. Cai, J. Wei, S. Qiu, S. Zhang, J. Pan, Z. Lin, and L. Wang for useful discussions. This work was supported by the National Natural Science Foundation of China under Contracts No. 12425505, No. 12235001, and No. U2230402. This work was also supported by the Fundamental and Interdisciplinary Disciplines Breakthrough Plan of the Ministry of Education of China-JYB2025XDXM204.
\end{acknowledgments}

\section*{Data Availability}
The code used to perform the analysis in this paper is publicly available on GitHub~\cite{song2025detailedbalancecode}. The data discussed in this paper are available on Hugging Face~\cite{song2026detailedbalanceddata}.

\nocite{Cai_2025,Cao:2025shc,Singhal2023,Song:2025pwy,Szymanski2023,chen2026molecular,hoover2024energybased,ideasearchfitter_repo,pmlb,qwen2025qwen25technicalreport}
\bibliography{cite}
\clearpage
\onecolumngrid
\begin{center}
{\Large\bfseries Supplemental Material}\\[0.5em]
{\large for ``Detailed Balance in Large-Language-Model-Driven Agents''}
\end{center}
\beginsupplement


\section{Equivalence of the Least Action Principle and the Variational Condition}
\label{app:least action}
This supplementary material proves the equivalence of the variational condition Eq.~(2) and the least action principle when $K(x)$ is a convex function. The proof is conducted in the discrete case.

Firstly, we prove that the variational condition is a necessary condition for the least action.
Assume that when the least action is satisfied, $V_{\mathcal{T}}$ does not satisfy the variational condition Eq.~(2), then there exists at least one state $f_0$ such that
\begin{align}
\frac{\delta \mathcal{S}}{\delta V(f_0)} \bigg|_{V=V_{\mathcal{T}}} \neq 0.
\label{eq:neq}
\end{align}

This implies that there must exist another potential function $V'_{\mathcal{T}}$ defined as
\begin{align}
  V'_{\mathcal{T}}(f) = V_{\mathcal{T}}(f) - \epsilon \delta_{f,f_0} \frac{\delta \mathcal{S}}{\delta V(f_0)} \bigg|_{V=V_{\mathcal{T}}},
\end{align}
where $\delta_{f,f_0}$ is the Kronecker delta, and $\epsilon$ is a sufficiently small constant. Expanding the action $\mathcal{S}$ around it yields
\begin{align}
\mathcal{S}[V'_{\mathcal{T}}] &= \mathcal{S}[V_{\mathcal{T}}] + \sum_f \frac{\delta \mathcal{S}}{\delta V(f)} \bigg|_{V=V_{\mathcal{T}}} (V'_{\mathcal{T}}(f) - V_{\mathcal{T}}(f)) + O(\epsilon^2) \nonumber\\
&= \mathcal{S}[V_{\mathcal{T}}] + \frac{\delta \mathcal{S}}{\delta V(f_0)} \bigg|_{V=V_{\mathcal{T}}} (V'_{\mathcal{T}}(f_0) - V_{\mathcal{T}}(f_0)) + O(\epsilon^2) \nonumber\\
&= \mathcal{S}[V_{\mathcal{T}}] - \epsilon \left( \frac{\delta \mathcal{S}}{\delta V(f_0)} \bigg|_{V=V_{\mathcal{T}}} \right)^2 + O(\epsilon^2).
\end{align}

By choosing a sufficiently small $\epsilon>0$, the first-order term dominates, and due to Eq.~\eqref{eq:neq}, we have
\begin{align}
\left( \frac{\delta \mathcal{S}}{\delta V(f_0)} \bigg|_{V=V_{\mathcal{T}}} \right)^2>0,
\end{align}
Thus,
\begin{align}
  \mathcal{S}[V'_{\mathcal{T}}] < \mathcal{S}[V_{\mathcal{T}}].
\end{align}
This contradicts the assumption that $V_{\mathcal{T}}$ is the least action, thus completing the proof.

Next, we prove that when $K(x)$ is a convex function, the potential function $V_{\mathcal{T}}$ satisfying the variational condition is necessarily a global minimum point of the action $\mathcal{S}$. This can be reduced to proving that the action $\mathcal{S}$ is a convex functional.

Let $V_1,V_2$ be any two potential functions, and $0\le \lambda \le 1$, then
\begin{align}
  &\mathcal{S}[\lambda V_1 + (1-\lambda)V_2] \nonumber\\
  =& \sum_{f,g} K(\lambda V_1(f) + (1-\lambda)V_2(f) - \lambda V_1(g) - (1-\lambda)V_2(g)) \mathcal{T}(g\gets f) \nonumber\\
  \le& \sum_{f,g} \lambda K(V_1(f) - V_1(g)) + (1-\lambda) K(V_2(f) - V_2(g)) \mathcal{T}(g\gets f) \nonumber\\
  =& \lambda \mathcal{S}[V_1] + (1-\lambda) \mathcal{S}[V_2],
\end{align}
where the inequality arises from the convexity of $K(x)$ and the positivity of $\mathcal{T}(g\gets f)$, thus completing the proof.

In summary, when $K(x)$ is a convex function, the variational condition Eq.~(2) and the principle of least action are equivalent.

\section{Implementation of agent}
\label{app:Ideasearch}

To test the detailed balance proposed in this Letter, LLMs need to be embedded into an agent framework to standardize their state space and state transitions. This Supplemental Material describes the implementation details of two completely different agent frameworks and supplements more experimental results, including the examination of detailed balance through the potential function for the Constrained Word Generation Agent using GPT-5 nano model and the direct examination of Eq.~(7) in the \texttt{IdeaSearchFitter} Agent. Finally, this Supplemental Material constructs an agent to demonstrate that detailed balance is not a trivial condition and does not necessarily hold when the state space cannot be given a specific meaning. For the Constrained Word Generation Agent using GPT-5 nano model and the \texttt{IdeaSearchFitter} Agent, there are almost no failed states, so we can directly take $N_0(f)=\sum_{g} N(g\gets f)$, as shown in Eq.~\eqref{eq:transition kernel ideasearch}.

The transition sampling counts, unique state counts, unique transition counts, and the number of states with sampling times greater than 1 for the two agents are shown in Table~\ref{tab:agent stats}. The code used to construct the Agents below can be found in the \href{https://github.com/SonnyNondegeneracy/detialed-balance-llm}{GitHub} repository.
\begin{table}[tbp]
    \caption{Database statistics for the two agents, including transition sampling counts, unique state counts, unique transition counts, and the number of states with sampling times greater than 1.}
\begin{ruledtabular}
\begin{tabular}{ccccc}
Agent & Transition Samples & Unique States & Unique Transitions & States with Samples $>1$ \\
\hline
\texttt{IdeaSearchFitter} & 50228 & 7484 & 21697 & 2551 \\
Conditioned Word Generation (GPT-5 nano) & 19968 & 645 & 9473 & 620 \\
\end{tabular}
\end{ruledtabular}
\label{tab:agent stats}
\end{table}

\subsection{Conditioned Word Generation Agent}
To examine the generative dynamics of the model, we constructed an agent $\mathcal{T}_{\text{real,I}}$ based on a conditioned word generation task. This agent generates a word through an LLM, requiring that the sum of the indices corresponding to all letters in the word equals 100 (for example, ATTITUDE, EXCELLENT). The state space of this agent consists of all words that satisfy this condition, and the large models used in state transitions are GPT-5 nano, Claude Sonnet 4.5, and Gemini 2.5 Flash, which read the context containing prompts and given states to generate new words that meet the condition. The specific implementation can be found in the \href{https://github.com/SonnyNondegeneracy/detialed-balance-llm}{GitHub} repository.

\begin{figure}
  \centering
  \includegraphics[width=0.45\linewidth]{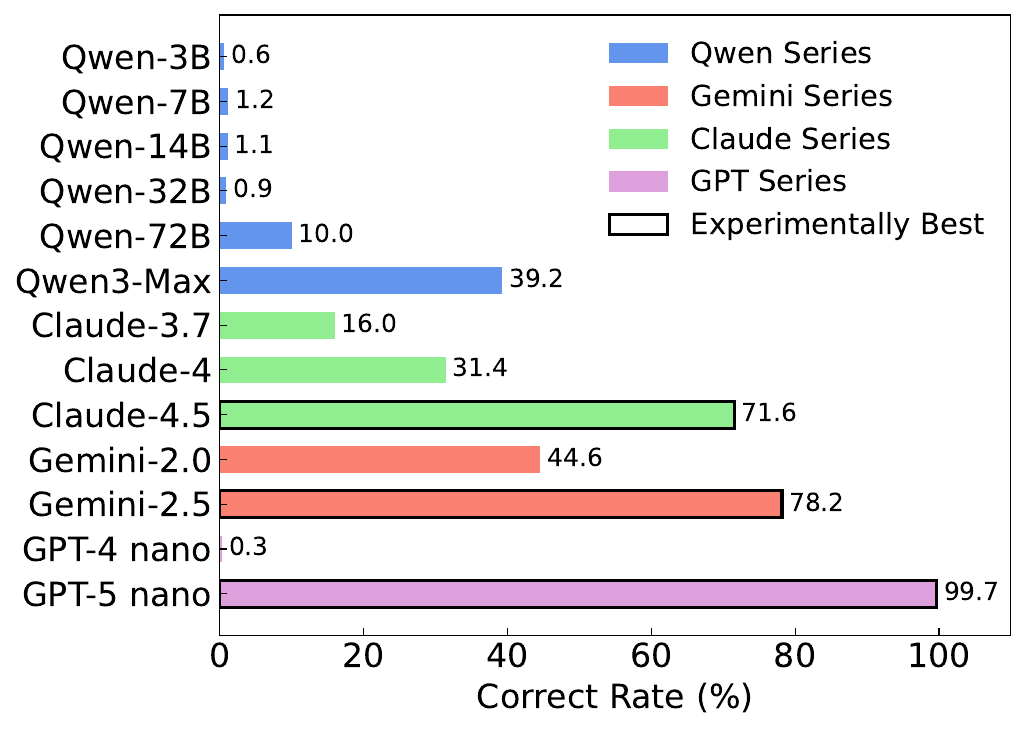}
  \caption{Comparison of $\text{Correct Rate}=1-\frac{\sum_f{N(\text{failed}\gets f)}}{\sum_f{N_0(f)}}$ for different LLMs in the conditioned word generation task. The Qwen series models are Qwen2.5-Instruct-3B, Qwen2.5-Instruct-7B, Qwen2.5-Instruct-14B, Qwen2.5-Instruct-32B, Qwen2.5-Instruct-72B and Qwen3-Max. The Claude series models are Claude Sonnet 3.7, Claude Sonnet 4, and Claude Sonnet 4.5. The Gemini series models are Gemini 2.0 Flash and Gemini 2.5 Flash. The GPT series models are GPT-4.1 nano and GPT-5 nano.}
  \label{fig:instruction following}
\end{figure}

As demonstrated in the main text, the state space is a unique characteristic of agents constructed by LLMs. We now show that the more mature the model, the stronger its instruction-following ability, thus making it more reasonable to study the generative dynamics of LLMs through the state space instead of the actual token sequence. To verify this, in addition to the three state-of-the-art models in the main text, we also compared the performance of the Qwen series, Claude Sonnet series, Gemini series, and GPT series models under the same conditioned word generation task~\cite{qwen2025qwen25technicalreport,claude4systemcard,geminimodelcard,gpt5systemcard}. For each model, we selected 100 different prompt words, generated 10 times for each word, and calculated the proportion of generated words that met the condition described in the prompt, i.e., the sum of letter indices equals 100, referred to as Correct Rate. The results are shown in Fig.~\ref{fig:instruction following}. It indicates that as the model capability or scale increases, the proportion of generated words that meet the condition improves. The most advanced models can almost completely follow instructions, making them most suitable for embedding into agents and using the state space to describe their generative dynamics. Therefore, we use these experimentally best-performing models to study the characteristics of the state space. It is reasonable to expect that future models will have stronger instruction-following capabilities, making the study of state space valuable. The role of training in shaping the potential function $V$ remains an open question; detailed balance is an empirical tendency observed across models of different scales and training procedures, but a theoretical derivation from first principles of training dynamics is left for future work.

\begin{figure}[tbp]
    \centering
    \includegraphics[width=0.5\linewidth]{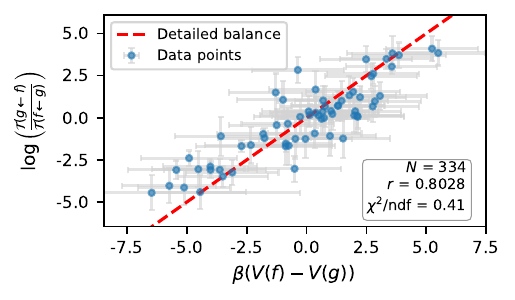}
    \caption{Test of the detailed balance condition for the Conditioned Word Generation Agent using GPT-5 nano model. The error estimation includes root mean square statistical errors, and the error of the potential function is estimated using the bootstrap method. $r$ is the Pearson correlation coefficient, and $\chi^2/\text{ndf}$ is the chi-squared per degree of freedom. Measurement includes state pairs with at least one measured transition between the two states. The agent's underlying potential function is consistent with the detailed balance condition, with systematic deviations from detailed balance observed at higher potential function values. Points with $|V_{\mathcal{T}}(f)-V_{\mathcal{T}}(g)|>\log{20000}$ are excluded. One-fifth of the data points are displayed for clarity.}
    \label{fig:word results}
\end{figure}

The implementation of the Conditioned Word Generation Agent shows two different behavioral patterns, with the Claude Sonnet 4.5 and Gemini 2.5 Flash models exhibiting significant convergence, while the GPT-5 nano model demonstrates broader exploration capabilities. Fig.~\ref{fig:word results} presents the results of testing the detailed balance condition for the Conditioned Word Generation Agent using the GPT-5 nano model.

\subsection{\texttt{IdeaSearchFitter} Agent}

To examine the performance of LLM generative dynamics in specific tasks, we constructed an agent $\mathcal{T}_{\text{real}}$ based on the symbol fitting task using $\texttt{IdeaSearchFitter}$~\cite{ideasearchfitter_repo}. The state space of this agent consists of strings represented as expression trees $f$, and state transitions are achieved by generating new expression trees through LLMs. The agent runs in expert mode 10 times to obtain the database used in the main text~\cite{song2025iteratedagentsymbolicregression}; specifically, ``example\_num'' is set to 1 to simplify the state space to numexpr strings, and ``auto\_polish'' is set to True to test with richer prompts. ``sample\_temperature'' and ``model\_sample\_temperature'' are set to 1000.0 to uniformly sample the state space. Each run searches the ``nikuradse\_2'' dataset from PMLB~\cite{pmlb} without early stopping conditions. The final dataset contains 50,228 state transitions, involving 21,697 unique transitions and 7,484 unique states, of which 2,551 states were sampled more than once.

Within this Agent, the method for estimating the transition kernel is:
\begin{align}
  \mathcal{T}(g\gets f) = 
  \begin{cases}
    \dfrac{N(g\gets f)}{\sum_{g'\neq f} N(g'\gets f)}, & \sum_{g'} N(g'\gets f) > 1 \text{ and } g\neq f,\\
    0, & \text{otherwise.}
  \end{cases}
  \label{eq:transition kernel ideasearch}
\end{align}
We excluded self-loop transitions ($f=g$) and states that were recorded only once ($\sum_{g'} N(g'\gets f) \le 1$) to make the estimation of the transition kernel more accurate. Here, $N(g\gets f)$ represents the number of transitions from state $f$ to state $g$.

\begin{figure}[tbp]
    \centering
    \includegraphics[width=0.9\linewidth]{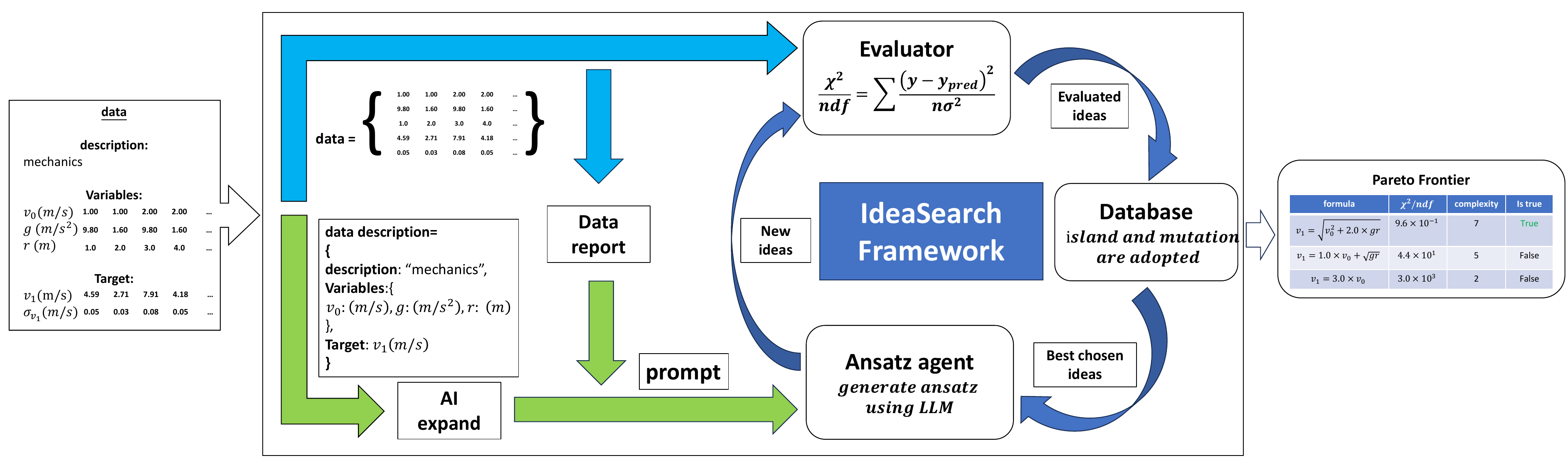}
    \caption{Architecture of the \texttt{IdeaSearchFitter} agent (reproduced from Ref.~\cite{song2025iteratedagentsymbolicregression}). Input data and variable descriptions are expanded into a structured prompt once at initialization, after which the system iterates: an ansatz agent (LLM) generates candidate expressions informed by the best-performing entries in a database, an evaluator scores them, and the results are stored in an island-based database for selection and mutation.}
    \label{fig:isfitter}
\end{figure}

The \texttt{IdeaSearchFitter} is a symbolic regression agent built on the IdeaSearch framework~\cite{ideasearch2025,song2025iteratedagentsymbolicregression}. A detailed description of the IdeaSearch framework can be found in Ref.~\cite{song2025iteratedagentsymbolicregression}; here we summarize the key aspects relevant to the detailed balance analysis. Its architecture is illustrated in Fig.~\ref{fig:isfitter}.

The workflow proceeds as follows: at initialization, the input data and variable descriptions are expanded by an AI module into a structured prompt. The system then iterates steps (1)--(3): (1)~an ansatz agent (LLM) reads this prompt together with the best-performing expressions from the current database and generates new candidate mathematical expressions; (2)~an evaluator scores each candidate against the data using $\chi^2/\text{ndf}$; (3)~the scored expressions are stored in a database organized into islands, where selection and mutation are applied before feeding the best examples back into the next generation cycle. Each generation step involves a rich, multi-part prompt containing the data description, prior best solutions, and evaluation feedback. The key hyperparameters are listed in Table~\ref{tab:isfitter_params}; the full configuration can be found in the \href{https://github.com/SonnyNondegeneracy/detialed-balance-llm}{GitHub} repository.

\begin{table}[tbp]
\centering
\caption{Key hyperparameters of the \texttt{IdeaSearchFitter} agent used in this work, taken from the \href{https://github.com/SonnyNondegeneracy/detialed-balance-llm}{GitHub} repository.}
\label{tab:isfitter_params}
\begin{ruledtabular}
\begin{tabular}{lc}
Parameter & Value \\
\hline
Parallel LLM calls per generation (\texttt{samplers\_num}) & 3 \\
Best examples shown in prompt (\texttt{examples\_num}) & 1 \\
Candidates per call (\texttt{generate\_num}) & 3 \\
Sampling temperature & 1000 \\
Number of LLM generators & 10 \\
Number of islands & 8 \\
Cycles per island & 30 \\
Interactions per cycle & 10 \\
\end{tabular}
\end{ruledtabular}
\end{table}

\begin{figure}[tbp]
    \centering
    \includegraphics[width=0.5\linewidth]{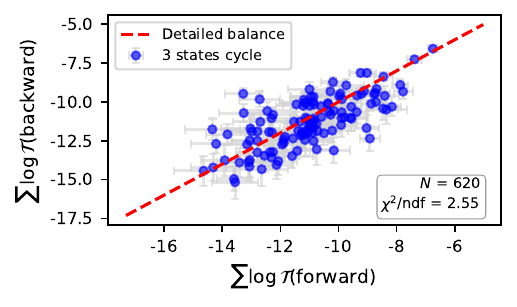}
    \caption{Test of detailed balance through closed paths in the state transition graph of the complex agent $\mathcal{T}_{\text{real}}$. Each point represents a triplet, with a total of 620 different triplets found in the experimental data, where each transition was detected at least twice. This indicates that within the error range, the sums of the logarithms of the forward and backward transition kernels are roughly equal, consistent with the detailed balance condition. To clearly display the figure, only 1/5 of the data points are shown.}
    \label{fig:closed loop}
\end{figure}

In the main text, we examined the detailed balance condition of the \texttt{IdeaSearchFitter} agent by directly comparing the differences in its potential function and the logarithm of the transition kernel ratios. To further test detailed balance, we also examined it through closed paths in its state transition graph. Specifically, we searched for all possible triplets $(f,g,h)$ in the experimental data such that transitions exist between each pair. For each triplet, we calculated the sums of the logarithms of the forward and backward transition kernels along the closed path, with the results shown in Fig.~\ref{fig:closed loop}. A total of 620 different triplets were found in the experimental data, indicating that within the error range, the sums of the logarithms of the forward and backward transition kernels are roughly equal, consistent with the detailed balance condition.

\begin{figure}[tbp]
    \centering
    \begin{minipage}{0.45\linewidth}
      \centering
      \includegraphics[width=0.9\linewidth]{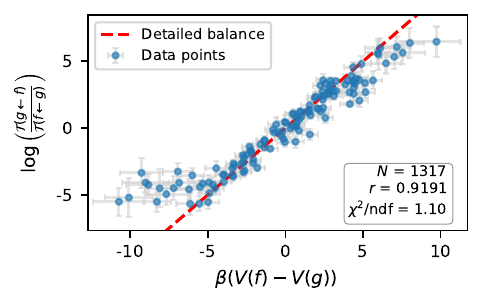}
      \end{minipage}\hfill
    \begin{minipage}{0.45\linewidth}
      \centering
      \includegraphics[width=0.9\linewidth]{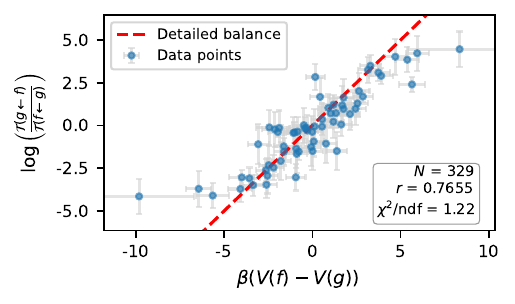}
    \end{minipage}
    \caption{Test of the detailed balance condition for both agents using $K(x) = \log{(1+e^{-x})}$, with all settings the same as in the main text. $r$ is the Pearson correlation coefficient, and $\chi^2/\text{ndf}$ is the chi-squared per degree of freedom. (a) Results for the \texttt{IdeaSearchFitter} agent. (b) Results for the Conditioned Word Generation Agent.}
    \label{fig:different K}
\end{figure}

Next, we demonstrate that even when changing the specific form of $K(x)$, as long as it satisfies Eq.~\eqref{eq:K condition}, the potential function distribution consistent with detailed balance can still be recovered through the principle of least action. Fig.~\ref{fig:different K} shows the results when using a common function form in transition dynamics, $K(x) = \log{(1+e^{-\beta x})}$. It can be seen that the potential function distribution is basically consistent with the results obtained using $K(x) = e^{-\beta x/2}$ in the main text, further supporting the reasonableness of the detailed balance condition.

\begin{figure}[tbp]
\centering
\begin{minipage}{0.45\linewidth}
  \centering
  \includegraphics[width=0.9\linewidth]{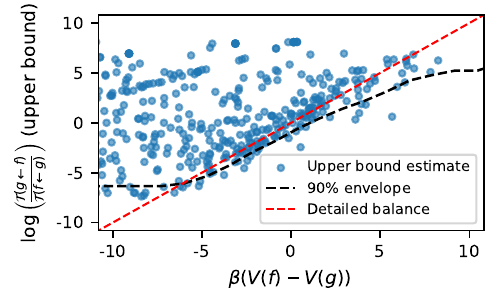}
\end{minipage}\hfill
\begin{minipage}{0.45\linewidth}
  \centering
  \includegraphics[width=0.9\linewidth]{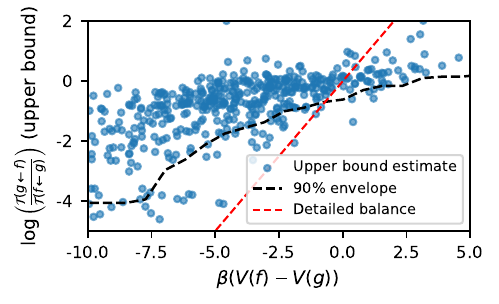}
\end{minipage}
\caption{For pairs in the \texttt{IdeaSearchFitter} and Conditioned Word Generation Agent where the transition $f\gets g$ was measured but the transition $g\gets f$ was not, we compare the differences in their potential functions and the logarithm of the transition kernel ratios. The figure shows points corresponding to 500 such pairs. The black dashed line represents the 90th percentile line. It is mostly above the red dashed line representing detailed balance, indicating that the inequality Eq.~\eqref{eq:exclude} is basically satisfied, with those points at larger $|\beta V_{\mathcal{T}}(f) - \beta V_{\mathcal{T}}(g)|$ possibly arising from systematic overestimation of the potential function differences. (a) Results for the \texttt{IdeaSearchFitter} agent with a total of 18,935 such pairs. (b) Results for the Conditioned Word Generation Agent with a total of 8,805 such pairs. From this figure, it can be directly seen that most of such transition pairs come from cases where $N(f\gets g)=1$.}
\label{fig:exclude detail}
\end{figure}

Finally, in order to further examine the consistency of detailed balance, we discuss such pairs where the transition $g \gets f$ was not measured, while the transition $f \gets g$ was measured. Using the same notation conventions as in the main text, we can estimate
\begin{align}
  \beta V_{\mathcal{T}}(f) - \beta V_{\mathcal{T}}(g) = \log{\frac{\mathcal{T}(g\gets f)}{\mathcal{T}(f\gets g)}} < \log{\frac{1/N(f)}{N(f\gets g)/N(g)}}.
  \label{eq:exclude}
\end{align}
Here, $N(f)$ represents the total number of transitions from state $f$. Fig.~\ref{fig:exclude detail} shows the comparison results for these pairs. It can be seen that these pairs basically satisfy the inequality Eq.~\eqref{eq:exclude}, which further supports the reasonableness of the detailed balance condition. It is worth emphasizing that since these pairs are not fully included in the convex optimization of the minimum action, the estimates of $|\beta V_{\mathcal{T}}(f) - \beta V_{\mathcal{T}}(g)|$ may sometimes be overestimated.

\subsection{Random Agent}
\begin{figure}[tbp]
\centering
\begin{minipage}{0.45\linewidth}
  \centering
  \includegraphics[width=0.9\linewidth]{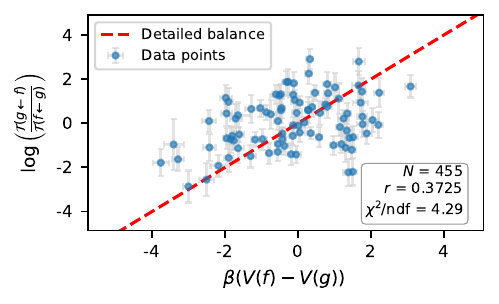}
\end{minipage}\hfill
\begin{minipage}{0.45\linewidth}
  \centering
  \includegraphics[width=0.9\linewidth]{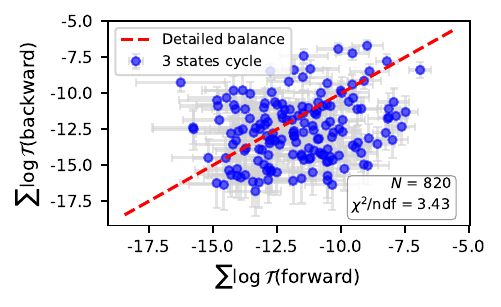}
\end{minipage}
\caption{Results of the detailed balance test for the random agent $\mathcal{T}_{\text{rand}}$. $r$ is the Pearson correlation coefficient, and $\chi^2/\text{ndf}$ is the chi-squared per degree of freedom. (a) Examination of the detailed balance condition by comparing the differences in potential functions and the logarithm of transition kernel ratios. (b) Examination of the detailed balance condition through closed paths in the state transition graph. It can be seen that the state transitions of this agent do not satisfy the detailed balance condition.}
\label{fig:random agent}
\end{figure}

To demonstrate that detailed balance is not a trivial condition, we constructed a random agent $\mathcal{T}_{\text{rand}}$. Specifically, the state space consists of random integer strings generated from a given seed, and the LLM is prompted to predict the next integer. This predicted integer is then used as a seed to generate a random string, which is considered the next state. We used the GPT-5 nano model to construct this agent and ran it 64,000 times, successfully measuring 14,705 transitions to obtain the state transition database. The detailed implementation can be found in the \href{https://github.com/SonnyNondegeneracy/detialed-balance-llm}{GitHub} repository. The results indicate that the state transitions of this agent do not satisfy the detailed balance condition, as shown in Fig.~\ref{fig:random agent}. This suggests that when the state space cannot be given a specific meaning that the model can understand, detailed balance does not necessarily hold. This situation may occur when the model's capability is insufficient to comprehend the semantics of the states, or when the states themselves involve excessive tool calls and external environment interactions, rendering it meaningless to view the agent as a system completing state transitions through LLM generation.

To provide a matched negative control for the additional Qwen-based semantic tasks described in the following subsection, we constructed a second random agent using the Qwen2.5-Instruct model family with the same pseudorandom-seed protocol. To obtain more robust statistics, we aggregated data from five model sizes (3B, 7B, 14B, 32B, and 72B), analogous to how the IdeaSearchFitter agent in the main text mixes outputs from multiple models. The combined dataset yields 66 unique states and 1,352 measured transitions. As shown in Table~\ref{tab:multi_task}, this Qwen-based random agent also fails to satisfy the detailed balance condition ($r=0.43$), consistent with the GPT-5 nano random agent above.

\subsection{Additional word-generation tasks}
\label{app:additional_tasks}

To test the generality of detailed balance beyond the letter-sum-equals-100 task, we designed three additional constrained word-generation tasks using Qwen2.5-72B-Instruct~\cite{qwen2025qwen25technicalreport}, each defining a different semantic state space.

\textbf{5-Letter-S.} The agent generates five-letter English words beginning with the letter ``S.'' Given a prompt word satisfying this constraint, the LLM is asked to produce another valid five-letter word starting with S. The state space consists of all such words recognized by the model. This task probes detailed balance in a state space where the constraint is purely lexical rather than arithmetic.

\textbf{Digit Sum$=$15.} The agent generates non-negative integers whose digits sum to 15 (e.g., 69, 78, 555). Given a prompt integer satisfying this constraint, the LLM produces another integer with the same digit-sum property. This task tests whether detailed balance holds when states are numerical rather than lexical.

\textbf{Sum$\div$10.} The agent generates English words whose letter-index sum (A$=$1, B$=$2, \ldots, Z$=$26) is divisible by 10. This is a relaxed variant of the letter-sum-equals-100 constraint used in the main text, yielding a larger and more diverse state space.

All three tasks follow the same agent protocol as the Conditioned Word Generation Agent: the LLM receives a single prompt state and generates one candidate state per call, with validity checked against the task constraint. The transition kernel is estimated using Eq.~\eqref{eq:transition kernel ideasearch}. The Qwen-based random agent described in the previous subsection serves as a matched negative control for these tasks.

\subsection{Multi-task test of detailed balance}
\label{app:multi_task}

To test the generality of detailed balance, we applied the same analysis pipeline to three additional word-generation tasks using Qwen2.5-72B-Instruct, together with the two reference models from the main text and random-agent negative controls. Table~\ref{tab:multi_task} summarizes the results.

\begin{table*}[tbp]
\caption{Multi-task test of detailed balance, with matched floor and ceiling controls. $r$ is the measured Pearson correlation between $\ln(\mathcal{T}(g\gets f)/\mathcal{T}(f\gets g))$ and $V(f)-V(g)$ for bidirectional pairs with both counts $>1$. $r_{\rm floor}$ and $r_{\rm ceil}$ are matched controls processed through the identical pipeline, quoted as medians over repeated randomizations; their construction and interpretation are described in the text.}
\label{tab:multi_task}
\begin{ruledtabular}
\begin{tabular}{lcccccc}
Task & States & Trans & $N_{\rm pairs}$ & $r$ & $r_{\rm floor}$ & $r_{\rm ceil}$ \\
\hline
\multicolumn{7}{l}{\emph{Semantic agents (Qwen2.5-72B):}}\\
5-Letter-S       & 206 & 4232 & 240 & 0.58 & 0.17 & 0.88 \\
Digit Sum$=$15   & 41  & 196  & 15  & 0.79 & 0.34 & 1.00 \\
Sum$\div$10      & 69  & 242  & 7   & 0.84 & 0.52 & 0.99 \\
\hline
\multicolumn{7}{l}{\emph{Reference models from main text:}}\\
GPT-5 nano word gen. & 645  & 9473  & 329  & 0.77 & 0.15 & 0.90 \\
IdeaSearchFitter      & 7484 & 21697 & 1317 & 0.92 & 0.00 & 0.97 \\
\hline
\multicolumn{7}{l}{\emph{Random-agent negative controls:}}\\
GPT-5 nano  & 64 & 1940 & 455 & 0.37 & 0.16 & 0.86 \\
Qwen-2.5    & 66 & 1352 & 172 & 0.43 & 0.15 & 0.95 \\
\end{tabular}
\end{ruledtabular}
\end{table*}

Agents with semantic state spaces yield $r\approx 0.58$--$0.92$, while random-agent controls yield $r\approx 0.37$--$0.43$, indicating that the tendency toward detailed balance is specific to LLM-driven generation with semantic content.

\begin{figure}[tbp]
    \centering
    \includegraphics[width=0.72\linewidth]{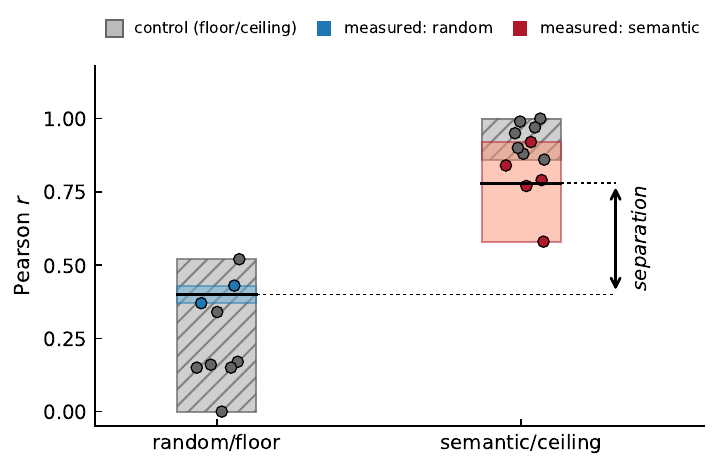}
    \caption{Two overlaid pairs of distributions of the consistency measure~$r$ for the tasks of Table~\ref{tab:multi_task}; the construction of the controls is described in the text. Left: the matched floor control $r_{\rm floor}$ (gray) overlaid with the measured random-agent controls (blue). Right: the matched ceiling control $r_{\rm ceil}$ (gray) overlaid with the measured semantic agents (red). Each box spans the min--max range across tasks within its group; the black bar marks the cross-task mean of the measured~$r$ in each group.}
    \label{fig:r_separation}
\end{figure}

We emphasize that $r$ is a continuous measure of consistency rather than a pass/fail criterion. To calibrate it, we construct two matched controls for each task (Table~\ref{tab:multi_task}), both processed through the identical fitting and correlation pipeline as the data. The floor $r_{\rm floor}$ randomizes the connectivity of the transition graph: each state keeps its observed outgoing transition counts, but the targets are redrawn uniformly at random among the states entering the fit, preserving the state set and the sampling scale; refitting the randomized counts yields the correlation produced by the fit itself once detailed balance is destroyed. The ceiling $r_{\rm ceil}$ redraws, for each state $f$, its $N_0(f)$ recorded transitions multinomially from a kernel consistent with detailed balance, $T(g\gets f)\propto\exp[V(f)-V(g)]$, restricted to the states that $f$ transitions to, with $V$ the task's fitted potential; refitting the resampled counts yields the $r$ attainable when detailed balance is imposed by construction at the same sampling scale. Built from the fitted potential, the ceiling is a reference set by the sampling scale, not an independent test of whether detailed balance holds. The measured $r$ lies between the two for every task ($r_{\rm floor}<r<r_{\rm ceil}$): it sits above the floor because the real counts retain structure consistent with detailed balance that the randomization removes, and below the ceiling by the part of the count structure that a kernel consistent with detailed balance does not reproduce, which is the finite deviation from detailed balance. Neither control is a strict bound; in particular, the floor lies above $0$ because the fit induces some correlation even for randomized counts. Random-agent controls lie closer to their floors than to their ceilings, whereas semantic agents lie closer to their ceilings (Table~\ref{tab:multi_task}, Fig.~\ref{fig:r_separation}). This separation is corroborated independently by the sums along closed paths (Fig.~3 of the main text and Fig.~\ref{fig:random agent}), which are consistent with zero for the semantic agent and inconsistent with zero for the random control.

\subsection{Ergodicity of the state transition graph}
\label{app:ergodicity}

For the high-temperature regime (GPT-5 nano, 645 unique words), we performed an ergodicity analysis of the full transition graph by computing its strongly connected components (SCCs). We find that 630 out of 645 states (97.7\%) belong to a single giant SCC, in which every state can be reached from every other state through a sequence of transitions. All four seed words (BUZZY, WIZARDS, AZURITE, EXCELLENT) lie within this component. The remaining 15 states (2.3\%) are transient: they were generated once but never served as a transition source, contributing less than 0.01\% of all transitions. Since nearly the entire state space forms a single ergodic class, the stationary distribution $\pi$ is unique and independent of the choice of initial seeds.

\subsection{Sensitivity of detailed balance to the sampling threshold}
\label{app:sensitivity}

To determine whether our conclusions are sensitive to the sampling distribution, we performed a sensitivity analysis on the IdeaSearchFitter data. We define $N_0(f)$ as the total number of times state $f$ appeared as a source state (i.e.\ the number of outgoing transitions from $f$), and $N_{\mathrm{pairs}}$ as the number of state pairs $(f,g)$ with at least one transition observed in each direction. We first fit the potential landscape $V(f)$ using all available samples, and then test detailed balance using Eq.~(5) of the main text only for state pairs $(f,g)$ with $N_0(f), N_0(g) \geq N_0^{\min}$, varying the threshold $N_0^{\min}$.

\begin{figure}[tbp]
    \centering
    \includegraphics[width=0.9\linewidth]{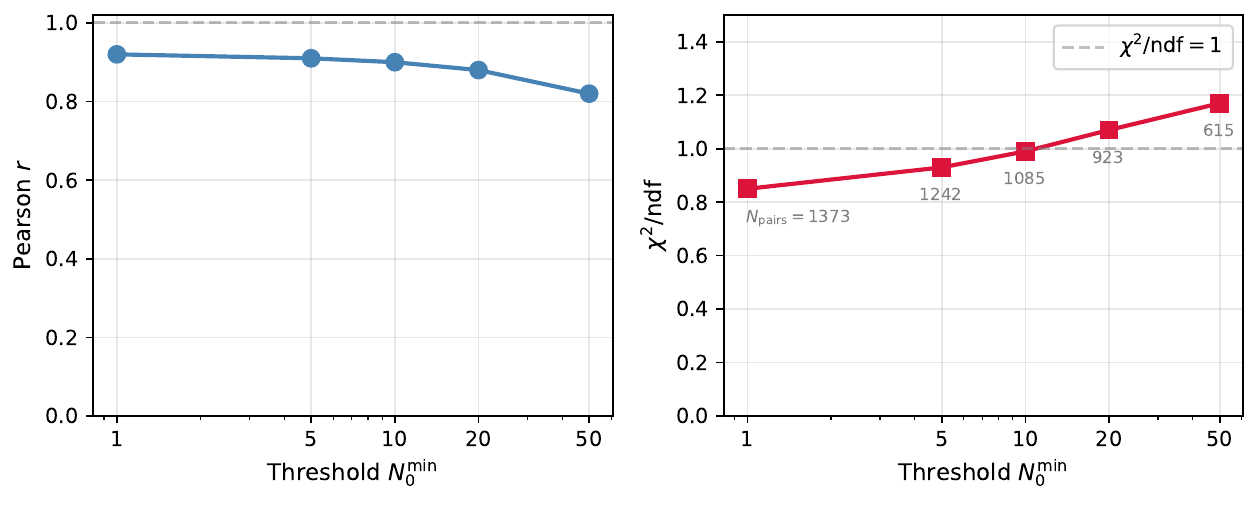}
    \caption{Sensitivity of the detailed-balance test to the minimum sampling threshold $N_0^{\min}$ for IdeaSearchFitter. $N_0(f)$ is the total number of outgoing transitions from state $f$, and $N_{\mathrm{pairs}}$ is the number of state pairs with bidirectional transitions passing the threshold. Left: Pearson $r$ between $\ln[\mathcal{T}(g\!\leftarrow\!f)/\mathcal{T}(f\!\leftarrow\!g)]$ and $\beta(V(f)-V(g))$ remains high ($r\simeq 0.82$--$0.92$) across all thresholds, with only a modest decline even when the dataset is restricted to the ${\sim}\,600$ best-sampled pairs. Right: $\chi^2/\mathrm{ndf}$ increases from ${\sim}\,0.85$ at $N_0^{\min}=1$, crossing unity near $N_0^{\min}=10$ and reaching ${\sim}\,1.2$ at $N_0^{\min}=50$; the rise reflects a slight mismatch between the globally fitted potential and the high-threshold subset, rather than a systematic violation of detailed balance.}
    \label{fig:sensitivity}
\end{figure}

The Pearson $r$ remains high across all thresholds (Fig.~\ref{fig:sensitivity}, left), declining only modestly from $0.92$ to $0.82$ even when the pair count drops from 1{,}373 to 615. The $\chi^2/\mathrm{ndf}$ rises from ${\sim}\,0.85$ at $N_0^{\min}=1$ through unity near $N_0^{\min}=10$ to ${\sim}\,1.2$ at $N_0^{\min}=50$ (Fig.~\ref{fig:sensitivity}, right). Because the potential $V$ is fitted once using all data, this mild overshoot is expected: at high thresholds the test set is a small, well-sampled subset whose local structure may differ slightly from the global fit. Overall, both metrics indicate that the tendency toward detailed balance is robust, and that the conclusions are not driven by poorly-sampled states.

\subsection{Comparison with PageRank}
\label{app:pagerank}

A natural alternative for estimating the potential function is to compute the stationary distribution $\pi$ via the power method (PageRank) and define $V_{\text{PR}}(f) \equiv -\ln\pi(f)$. We tested this approach using damped power iteration with damping parameter $\alpha = 0.99$.

\begin{figure}[tbp]
    \centering
    \includegraphics[width=0.5\linewidth]{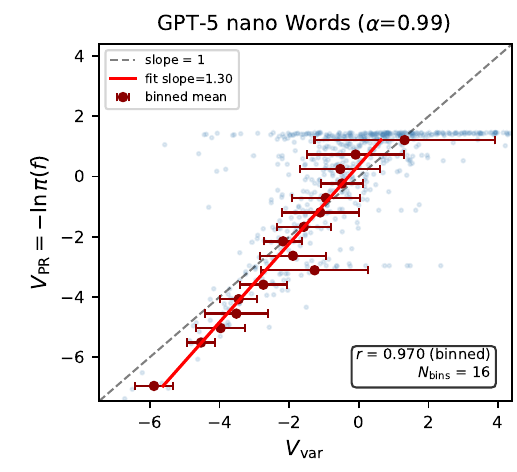}
    \caption{Comparison of PageRank potential $V_{\text{PR}} = -\ln\pi(f)$ and variational potential $V_{\text{var}}$ for the GPT-5 nano word task with damping factor $\alpha=0.99$. Light blue points are per-state values; dark red points are binned means with $1\sigma$ error bars. The dashed line has slope~1; the solid line is a linear fit (slope $=1.30$). The binned Pearson correlation is $r = 0.97$.}
    \label{fig:pagerank}
\end{figure}

The two estimates are strongly consistent: for the GPT-5 nano word task the binned Pearson correlation is $r = 0.97$ (Fig.~\ref{fig:pagerank}), indicating that both procedures recover the same underlying energy landscape. However, we do not adopt PageRank as the primary method, because it already presupposes the existence of a stationary distribution with a Boltzmann form when estimating the potential, and thus cannot serve as an independent test of whether detailed balance holds. The variational approach, by contrast, estimates $V$ directly from transition probability ratios without assuming any specific functional form for the stationary distribution.

\section{Low-temperature regime}
\label{app:special}

This Supplemental Material proves that the detailed balance condition Eq.~(5) is a sufficient condition for the variational principle. It then introduces some techniques for analytically calculating the minimum value of the action, based on which the potential function distributions of the Claude Sonnet 4.5 are analyzed.

\subsection{Detailed balance condition is a sufficient condition for the variational principle}
\begin{proof}
Assume that the agent $\mathcal{T}$ satisfies the detailed balance condition Eq.~(5), then for any state pair $(f,g)$, we have
\begin{align}
  \mathcal{T}(f\gets g) = \mathcal{T}(g\gets f)e^{-\beta (V_{\mathcal{T}}(f) - V_{\mathcal{T}}(g))}.
\end{align}
Therefore, substituting this relation into the equilibrium condition Eq.~(3), we obtain
\begin{align}
  \frac{1}{2}\sum_{f,g} \left[ K'(V_{\mathcal{T}}(f) - V_{\mathcal{T}}(g)) - K'(V_{\mathcal{T}}(g) - V_{\mathcal{T}}(f)) e^{-\beta (V_{\mathcal{T}}(f) - V_{\mathcal{T}}(g))} \right]\mathcal{T}(g\gets f) = 0.
\end{align}

Substituting the derivative of $K(x)$ into the above equation, the result is naturally satisfied.
\end{proof}

It's worth noting that any $K(x)$ function satisfying
\begin{align}
  K'(x) - K'(-x)e^{-\beta x} = 0,
  \label{eq:K condition}
\end{align}
makes the detailed balance condition a sufficient condition for the variational principle. There are no specific requirements for the form of $K(x)$ here.

\subsection{Analytically calculating the minimum action value}
Taking the Claude Sonnet 4.5 model as an example, we demonstrate how to analytically calculate the minimum action value to analyze the distribution of the potential function. Since only 5 different prompt words are involved, a relatively accurate estimation method is $N_0(f)=\sum{N(g\gets f)}$, thus estimating the transition kernel $\mathcal{T}(g\gets f)$ as
\begin{align}
    \mathcal{T}(g\gets f) \approx \frac{N(g\gets f)}{N_0(f)} \approx \frac{N(g\gets f)}{\sum_{g'} N(g'\gets f)}.
\end{align}
where $N(g\gets f)$ is the number of transitions from state $f$ to state $g$ as shown in the table~\ref{tab:count claude}.
We excluded self-loop transitions ($f=g$) and states that were recorded only once ($\sum_{g'} N(g'\gets f) \le 1$). We also recorded the unsuccessful generation cases for more detailed discussion, i.e., $\mathcal{T}(\text{failed}\gets f)=1-\sum_g' \mathcal{T}(g'\gets f)$. Here, $N(g\gets f)$ represents the number of transitions from state $f$ to state $g$.

\begin{table}[tbp]
\centering
\begin{tabular}{c|ccccccc}
\hline
From $\backslash$ To & ATT. & TUR. & PER. & PRO. & BUZ. & DIS. & failed \\
\hline
ATT. & 0 & 0 & 31 & 27 & 0 & 4 &3480\\
TUR. & 3521 & 0 & 0 & 0 & 0 & 0 & 74\\
PER. & 3500 & 0 & 0 & 0 & 0 & 0 & 0\\
PRO. & 3401 & 0 & 0 & 0 & 0 & 0 & 0\\
BUZ. & 2055 & 1254 & 0 & 0 & 0 & 0 & 218\\
DIS. & 2435 & 0 & 0 & 0 & 0 & 0 & 0\\
\hline
\end{tabular}
\caption{Transition kernel $\mathcal{T}(g\gets f)$ for the Claude Sonnet 4.5 model, where ATT., TUR., PER., PRO., BUZ., and DIS. represent the states ATTITUDE, TURKEY, PERSONAL, PROBLEM, BUZZY, and DISCIPLINE respectively. Each row represents the number of transitions starting from state $f$, and each column represents the number of transitions to state $g$. The reason for ``failed'' is that some transitions are rejected because they are words but the sum of letters is not 100, or the generated word is still the prompt word, especially for transitions starting from the ATTITUDE state.}
\label{tab:count claude}
\end{table}

Now, setting the zero point of the potential function as $V_{\mathcal{T}}(\text{ATT.})=0$, for the state PERSONAL, which has transitions only with the ATTITUDE state, the equilibrium condition Eq.~(3) degenerates into the detailed balance condition, yielding
\begin{align}
  \beta V_{\mathcal{T}}(\text{PER.}) \approx -\log{\frac{\mathcal{T}(\text{PER.}\gets \text{ATT.})}{\mathcal{T}(\text{ATT.}\gets \text{PER.})}} = -\log{\frac{31/3542}{3500/3500}} \approx 4.7.
\end{align}
Similarly, for the state PROBLEM, we have $V_{\mathcal{T}}(\text{PRO.}) \approx 4.8$ and $V_{\mathcal{T}}(\text{DIS.}) \approx 6.4$.
For the state BUZZY, there are no transitions into it, meaning that its potential function value cannot be estimated; we can only infer that its value is large, likely much greater than $\sim \log(20000)\sim 10$, which should be the maximum range for accurate measurement of the potential function.

For the state TURKEY, there are two transitions: one from ATTITUDE to TURKEY and another from BUZZY to TURKEY. Since it has been assumed that $V_{\mathcal{T}}(\text{BUZZY})\to \infty$, the equilibrium condition Eq.~(3) simplifies to
\begin{align}
  K'(\beta V_{\mathcal{T}}(\text{TUR.}) - \beta V_{\mathcal{T}}(\text{ATT.}))\mathcal{T}(\text{ATT.}\gets \text{TUR.}) = K'(\beta V_{\mathcal{T}}(\text{BUZ.}) - \beta V_{\mathcal{T}}(\text{TUR.}))\mathcal{T}(\text{TUR.}\gets \text{BUZ.}),
\end{align}
thus obtaining
\begin{align}
    \beta V_{\mathcal{T}}(\text{TUR.}) \to \infty.
\end{align}
In summary, the potential function of the Claude Sonnet 4.5 model is approximately
\begin{align}
  \beta V_{\mathcal{T}}(f) \approx \begin{cases}
    0, & f=\text{ATT.},\\
    4.7, & f=\text{PER.},\\
    4.8, & f=\text{PRO.},\\
    6.4, & f=\text{DIS.},\\
    \sim \infty, & f=\text{BUZ. or TUR.},\\
  \end{cases}.
\end{align}
This is consistent with the results in Fig.~2. It is worth noting that in this case, the potential function is almost self-consistently derived directly from the results of the detailed balance condition.

\begin{figure}
    \centering
    \includegraphics[width=0.5\linewidth]{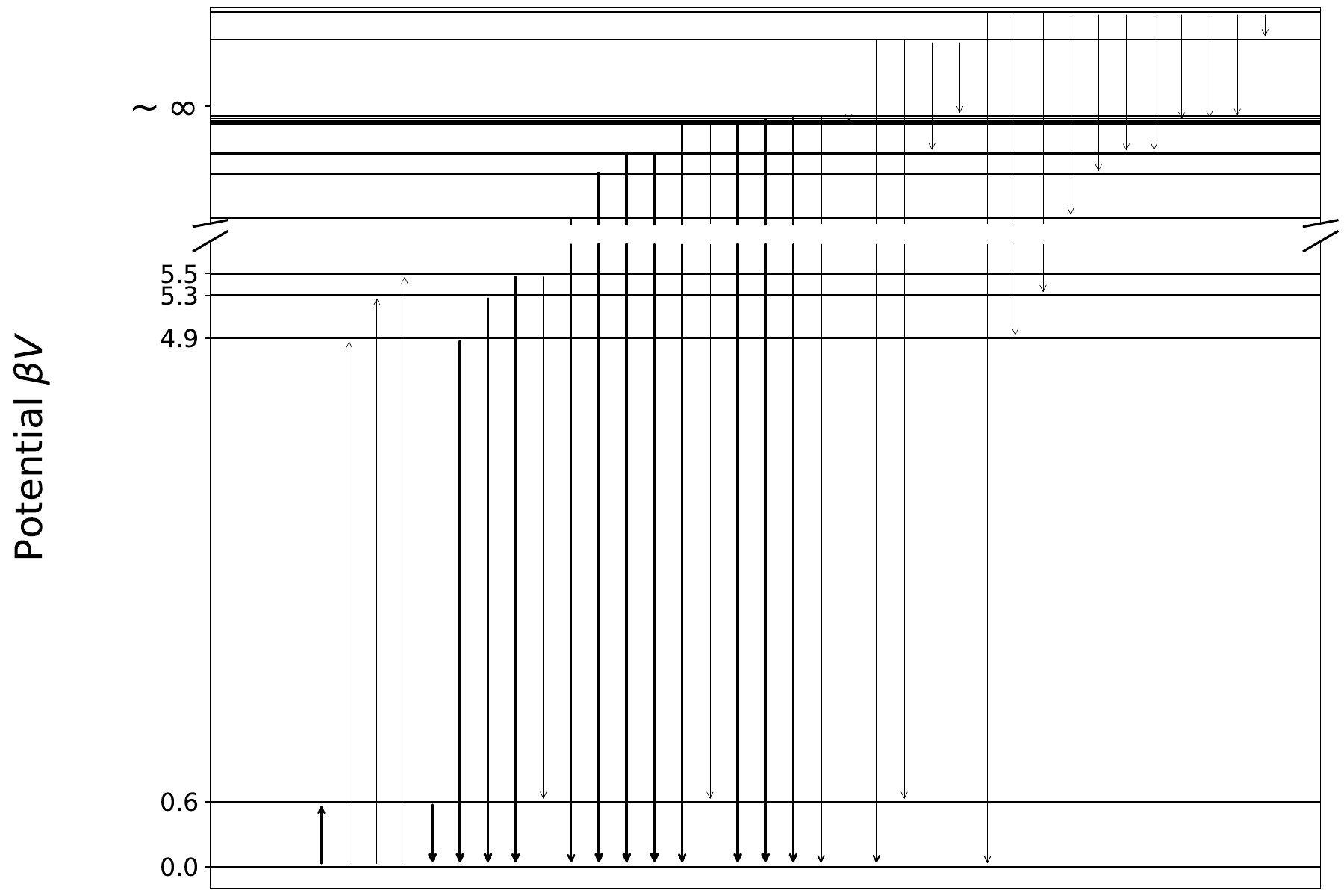}
    \caption{Transition process of the Gemini 2.5 Flash model in the prompt word state space, sorted by the potential function $V_{\mathcal{T}}$. Transitions tend to move towards states with lower potential functions. States with $\beta V(f)\gg log(20000)\sim 10$ are those where the equilibrium condition cannot be strictly satisfied. Thick lines represent high-frequency transitions, while thin lines represent low-frequency transitions. Each horizontal line represents a state, arranged in order of increasing potential function.}
    \label{fig:transition_matrix_gemimi}
\end{figure}

Similarly, analyzing the Gemini 2.5 Flash model yields the transition map shown in Fig.~\ref{fig:transition_matrix_gemimi}.

The potential function of Gemini 2.5 Flash is approximately
\begin{align}
    \beta V_{\mathcal{T}}(f) \approx \begin{cases}
        0, & f=\text{ATTITUDE},\\
        0.6, & f=\text{DISCIPLINE},\\
        4.9, & f=\text{BLISSFUL},\\
        5.3, & f=\text{EXCELLENT},\\
        5.5, & f=\text{CONGRESS},\\
        \sim \infty, & f=\text{others},
    \end{cases}.
\end{align}

\subsection{Initial condition convergence}
\label{app:convergence}

To demonstrate that the convergence of Claude Sonnet 4.5 and Gemini 2.5 Flash to a small number of states is a genuine dynamical property rather than an artifact of the choice of initial seed words, we performed a chain experiment. We randomly selected 100 diverse words satisfying the letter-sum-equals-100 constraint as starting points, including common English words (e.g., TURKEY, PERSONAL), rare words (e.g., BUZZY, KWANZAS), and nonsensical strings (e.g., ZZXX, TTTTT) that carry no semantic meaning but satisfy the numerical constraint. For each starting word, we ran a Markov chain of 10 steps using the same prompt template as in the main experiments.

\begin{table}[tbp]
\centering
\caption{Chain experiment: 100 independent Markov chains from diverse starting words, each running 10 steps. Starting words include common English words, rare words, and nonsensical strings satisfying the letter-sum-equals-100 constraint.}
\label{tab:chain_experiment}
\begin{ruledtabular}
\begin{tabular}{lcc}
Metric & Claude Sonnet 4.5 & Gemini 2.5 Flash \\
\hline
Chains converging to ATTITUDE & 89/100 (89\%) & 99/100 (99\%) \\
Mean steps to first ATTITUDE & 2.3 & 1.0 \\
Median steps to first ATTITUDE & 2 & 1 \\
ATTITUDE fraction of valid outputs & 68.8\% & 54.1\% \\
Second most common word & TURKEY (16.6\%) & DISCIPLINE (45.9\%) \\
Overall validity rate & 15.7\% & 75.3\% \\
\end{tabular}
\end{ruledtabular}
\end{table}

Regardless of the starting word, both models converge to ATTITUDE within 1--2 steps. This demonstrates that the convergence is a genuine property of the potential landscape rather than an artifact of the initial seed selection.

\section{Detailed discussion on the meaning of action}
\label{app:low tem}
The generative dynamics of LLMs are often highly directional. In the main text, we pointed out that the strength of this directionality can be measured by the size of the minimum action. In this Supplemental Material, we show that the action actually provides a method for estimating how the state density in the LLM-generated state space varies with the potential, and we use Majority Voting as an example to show that although the measurement of the potential function must be performed through the agent, the distribution of the potential function is not sensitive to the specific design of the agent to the extent that detailed balance holds.

\begin{figure}[tbp]
\centering
\begin{minipage}{0.45\linewidth}
  \centering
  \includegraphics[width=0.9\linewidth]{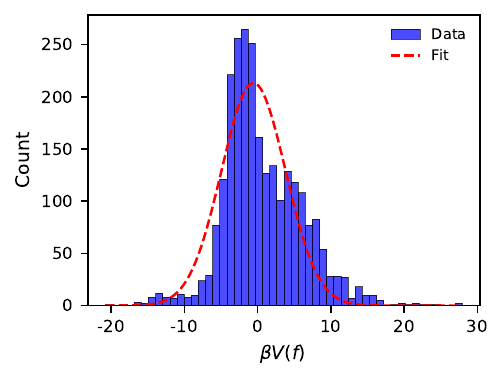}
\end{minipage}\hfill
\begin{minipage}{0.45\linewidth}
  \centering
  \includegraphics[width=0.9\linewidth]{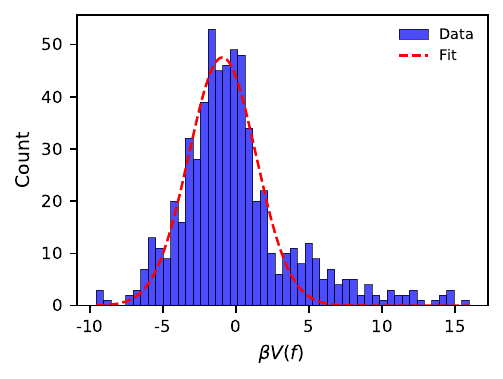}
\end{minipage}
\caption{(a) Distribution of potential functions for all states sampled at least twice in the \texttt{IdeaSearchFitter} agent. It exhibits a localized structure and can be fitted with a Gaussian distribution $\mathcal{N}(\mu=-0.56, \sigma=4.37)$. (b) Distribution of potential functions for all states sampled at least twice in the Conditioned Word Generation Agent. It exhibits a localized structure and can be fitted with a Gaussian distribution $\mathcal{N}(\mu=-0.93, \sigma=2.30)$.}
\label{fig:state distribution}
\end{figure}

Taking the \texttt{IdeaSearchFitter} agent and the Conditioned Word Generation Agent constructed with GPT-5 nano as examples, we measured the distribution of potential functions for all states in the database that were sampled at least twice, as shown in Fig.~\ref{fig:state distribution}. The distributions of both agents exhibit significant localized structures, indicating that the state density of the agents may be localized; in other words, the long-term behavior of these agents may be insensitive to the sampling temperature in the state space.
The distribution can be fitted with a Gaussian distribution $\mathcal{N}(\mu, \sigma)$. The relatively high standard deviation indicates that this potential function distribution is wide, within which the agent can exhibit significant directionality.

To quantify the relationship between action and state distribution, we assume that for a typical transition $g\gets f$, it satisfies $\beta V(f) - \beta V(g) \gg 1$. Considering detailed balance, we can approximately write
\begin{align}
  \mathcal{S} &\approx \int\int_{f,g\text{ for }V(f)<V(g)}{{ 2 \exp(\beta (V(f) - V(g))) \mathcal{T}(g\gets f)Df}Dg}.
\end{align}
If we only consider the scaling introduced by detailed balance, without considering the more specific structure of the state transition kernel, we can assume that (taking $\beta=1$ to simplify the notation):
\begin{align}
    \mathcal{S} &\sim \int\int_{V_f<V_g}{{ \frac{2}{2\pi \sigma^2} \exp{\left[(V_f-V_g)-\frac{V_f^2}{2\sigma^2}-\frac{V_g^2}{2\sigma^2}\right]} dV_f}dV_g}\nonumber\\
    & = \int_{-\infty}^{+\infty} dV_f \int_{V_f}^{+\infty} dV_g {\frac{2}{2\pi \sigma^2} \exp{\left[(V_f-V_g)-\frac{V_f^2}{2\sigma^2}-\frac{V_g^2}{2\sigma^2}\right]}}.
\end{align}
By making the variable substitutions $u=V_f+V_g$ and $v=V_g-V_f$, we can obtain
\begin{align}
    \mathcal{S} & \sim \int_{0}^{+\infty} dv \int_{-\infty}^{+\infty} du {\frac{2}{2\pi \sigma^2} \exp{\left[-v -\frac{u^2+v^2}{4\sigma^2}\right]}}\nonumber\\
    & = \int_{0}^{+\infty} dv {\frac{2}{\sqrt{\pi}\sigma} \exp{\left[-v -\frac{v^2}{4\sigma^2}\right]}}\nonumber\\
    & = 2 e^{\sigma^2} \text{erfc}(\sigma),
\end{align}
where erfc is the complementary error function. For large $\sigma$, we can approximately write $\mathcal{S} \approx \frac{\mathcal{S}_{\sigma=0}}{\sigma \sqrt{\pi}}$, indicating that the size of the action is inversely proportional to the standard deviation of the state density. Note that the normalization should be chosen to match the measurement as $\mathcal{S}_{\sigma=0}=K(0)$. The comparison of the expected minimum action size through the potential function and the actual minimized action for the two agents is shown in Table~\ref{tab:action comparison}.

\begin{table}[tbp]
\caption{Comparison of expected minimum action size through the potential function and actual minimized action for two agents.}
\begin{ruledtabular}
\begin{tabular}{cccc}
Agent & $\sigma$ & Expected min. action & Actual min. action \\
\hline
\texttt{IdeaSearchFitter} & 4.38 & 0.129 & 0.150 \\
Conditioned Word Generation(GPT-5 nano) & 2.30 & 0.245 & 0.195 \\
\end{tabular}
\end{ruledtabular}
\label{tab:action comparison}
\end{table}

This suggests that the size of the action estimates the characteristic energy scale in the agent's transitions, with smaller actions indicating that the agent's transition dynamics are more directional, while larger actions suggest that it is difficult for the agent's transitions to exhibit a clear directionality.

It is worth noting that measuring action is a more efficient method compared to directly measuring the directional distribution of the entire state space. In practical applications, the agent's transitions can be sampled through limited measurements to estimate the characteristic energy scale corresponding to this generative dynamics, thereby helping to improve the design of agent tasks more efficiently. For example, as shown in the main text, it can be seen that when poorer fits are sampled by \texttt{IdeaSearchFitter}, hyperparameters should be controlled to reduce the action and enhance the directionality of the agent's generation. While after reaching a better fit, the directionality of the internal generative dynamics of the LLM no longer aligns with that required by the optimization function, and hyperparameters should be controlled to increase the action, allowing the agent to serve as a mutation core~\cite{funsearch,song2025iteratedagentsymbolicregression} to explore the state space more efficiently without directional constraints.

We next illustrate that when detailed balance is satisfied, the design of the agent often only changes the scale of the potential function rather than its distribution. Therefore, the size of the action may serve as a universal metric for agent design. First, we assume that before each agent transition, instead of directly generating and extracting a new state, $M$ candidate states are generated, and then the state that appears more than $n>M/2$ times among the candidate states is selected as the new state (if no state meets this condition, the transition is rejected).
Under this design, assuming the original agent transition kernel is $\mathcal{T}(g\gets f)$, the new transition kernel can be written as
\begin{align}
  \mathcal{T}'(g\gets f) &= \sum_{k=n}^{M} \binom{M}{k} [\mathcal{T}(g\gets f)]^k [1-\mathcal{T}(g\gets f)]^{M-k}\nonumber\\
  &= I_{\mathcal{T}(g\gets f)}(n,M-n+1),
\end{align}

where $I_{x}(a,b)$ is the regularized incomplete beta function. Assuming $\mathcal{T}$ is not very large, we have
\begin{align}
  \frac{\mathcal{T}'(g\gets f)}{\mathcal{T}'(f\gets g)} & = \frac{I_{\mathcal{T}(g\gets f)}(n,M-n+1)}{I_{\mathcal{T}(f\gets g)}(n,M-n+1)} \approx \left(\frac{\mathcal{T}(g\gets f)}{\mathcal{T}(f\gets g)}\right)^n.
  \label{eq:beta}
\end{align}

\begin{figure}
    \centering
    \includegraphics[width=0.9\linewidth]{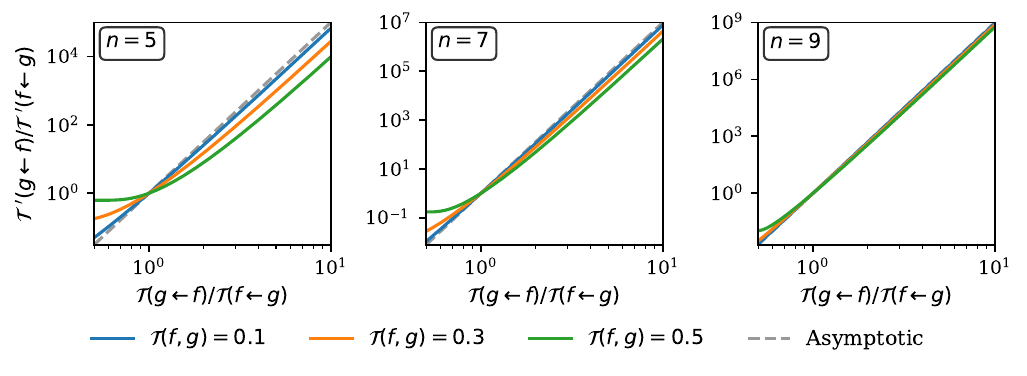}
    \caption{Numerical comparison of both sides of Eq.~\eqref{eq:beta}. The comparison is performed for $M=10$, with different colors representing different values of $\mathcal{T}(g\gets f)$. The three subplots represent the cases of $n=5,7,9$, respectively.}
    \label{fig:beta function}
\end{figure}

Numerical comparison is shown in Fig.~\ref{fig:beta function}. It can be seen that when $\mathcal{T}(g\gets f)$ is small, Eq.~\eqref{eq:beta} basically holds. This means that the majority-vote mechanism rescales the potential function by a factor of $n$:
\begin{align}
  V_{\mathcal{T}'} \approx n\, V_{\mathcal{T}}.
  \label{eq:majority_vote_rescaling}
\end{align}
Choosing the agreement threshold $n$ is therefore equivalent to adjusting the effective inverse temperature $\beta$: a higher $n$ sharpens the potential landscape ($\beta_{\text{eff}} \to n\beta$) and enhances directionality, while a lower $n$ flattens it and promotes exploration. This provides a tunable temperature knob at the agent level without modifying the underlying potential landscape, suggesting that the specific design of the agent has little effect on the distribution of the potential function. For this issue, we also realize that $\mathcal{S} \sim \frac{K(0)}{\sqrt{\pi}\sigma} \sim \frac{1}{n}$, indicating that by increasing the selection threshold $n$, the action can be effectively reduced, thereby enhancing the directionality of the agent's generative dynamics, while increasing $M$ does not significantly affect the size of the action but can improve sampling success rates by increasing the sampling budget when the transition kernel is small~\cite{meyerson2025solvingmillionstepllmtask}.

Generally speaking, an agent performing tasks within the distribution (e.g., in fields such as healthcare, experiments, etc.~\cite{Singhal2023,boiko2023autonomous,Szymanski2023}) should be designed to have a lower action, while an agent performing tasks outside the distribution (e.g., in fields such as exploring the frontiers of mathematics and theoretical physics~\cite{funsearch,Song:2025pwy,Cao:2025shc,alphaevolve,song2025iteratedagentsymbolicregression}) should be designed to have a higher action.

\section{Discovery of Potential Function using IdeaSearch}
\label{app:search}
In this Supplemental Material, we describe how to configure IdeaSearch and present the form and meaning of the best potential function discovered.
IdeaSearch is an automatic program search method based on LLMs~\cite{ideasearch2025,song2025iteratedagentsymbolicregression}, which can iteratively evaluate and optimize programs to solve complex problems by combining LLMs and evolutionary algorithms. We use IdeaSearch to search for the expression form of the potential function. Specifically, we represent the expression of the potential function as a combination of a predefined list of functions and operators, and the goal of IdeaSearch is to find an expression that maps this list to a floating-point number to minimize the corresponding action. IdeaSearch is configured with 16 different models to search for the target potential function. After running for 4000 rounds, the parameters in the best potential function were manually replaced and optimized using a random descent algorithm to minimize the action. The potential function discovered using IdeaSearch and the specific configuration parameters, evaluation, and running scripts of IdeaSearch can be found in the \href{https://github.com/SonnyNondegeneracy/detialed-balance-llm}{GitHub} repository.

The best potential function assigns a potential value by tokenizing the input mathematical function string and then evaluating its structure and complexity, normalizing the value to the range [-1, 1]. The function considers not only the syntactic integrity of the input string but also extracts various features, including function usage, parameter structure, and specific substructures. These features are combined to form the final expression. It is important to emphasize that even though the structure of the potential function discovered by IdeaSearch does not capture the non-commutativity of the potential function with respect to strings, it can still be used to estimate the directionality of the agent's transitions. This indicates that the directionality of agent transitions arising from LLM generation can be observed at different levels of coarse-graining, such as string level or expression level.

\section{Relation to Concept Networks and Energy-Based Transformers}
\label{app:related_work}

Two recent lines of work provide complementary perspectives on the structure of LLM generative dynamics. They also help clarify how detailed balance coexists with the manifest directionality of LLM reasoning: the trajectory-level drift on $V(f)$ and the state-pair ratio constraint Eq.~(5) of the main text operate at different resolutions, analogous to net molecular flow coexisting with microscopic reversibility in statistical physics~\cite{schroeder2020introduction,Cai_2025,chen2026molecular}.

\textbf{Concept Network Model (CoNet).} Hu \emph{et al.}~\cite{hu2025conceptnetwork}, together with the companion Learning-at-Criticality study by Cai \emph{et al.}~\cite{Cai_2025}, model the RLVR training process as dynamics on a directed random-regular graph (``concept network'') with $N=800$ nodes and average degree $\langle k\rangle\!\approx\!2$. The correspondence between CoNet simulations and real LLM behavior is validated at the level of global training observables: two-stage reward curves, V-shaped response-length curves, and the timing of catastrophic forgetting. CoNet's nodes are abstract integer labels; they do not correspond to tokens, hidden states, or semantic chunks, and neither work provides a projection from LLM outputs onto these nodes. Because CoNet is designed to reproduce macroscopic training curves rather than to resolve individual state-to-state transitions, it operates at a coarser granularity than our framework: it has no analogue of the pairwise transition kernel $\mathcal{T}(g\!\leftarrow\!f)$ or of the potential difference $V(f)-V(g)$ between identified states. Our detailed balance analysis, by contrast, works at the level of semantically defined state pairs within the generative (not training) dynamics, providing a thermodynamic constraint that is invisible at the topological resolution of CoNet. The two perspectives are therefore complementary: CoNet captures the global topological organization of the reasoning graph during training, while our framework characterizes the local thermodynamic structure of the state transitions during generation.

\textbf{Energy-based models and Energy-Based Transformers.} The thermodynamic analogy developed in this work naturally raises a more microscopic question: what, if anything, plays the role of the Hamiltonian at the token level? For standard autoregressive Transformers no explicit energy is defined; tokens are sampled from a learned conditional distribution, and any notion of ``energy'' must be reconstructed after the fact (as we do with $V$). Energy-based models (EBMs) provide a general framework in which a scalar energy function $E_\theta$ directly parametrizes unnormalized probabilities. Energy-Based Transformers (EBTs)~\cite{hoover2024energybased} are a recent instantiation of this framework: they equip a Transformer backbone with a per-example energy head, defining $E_\theta(x,\hat y)\in\mathbb{R}$ for each (context, candidate-prediction) pair. The standard EBM prescription then gives a joint distribution
\begin{align}
  p_\theta(x,\hat y) \;\propto\; e^{-E_\theta(x,\hat y)},
  \label{eq:ebt_p}
\end{align}
though in practice the partition function is intractable and the model works entirely with unnormalized energies, generating predictions by gradient-descent energy minimization rather than by sampling from Eq.~\eqref{eq:ebt_p}. Nevertheless, Eq.~\eqref{eq:ebt_p} provides a useful theoretical starting point for understanding what $V(f)$ represents. Note that Eq.~\eqref{eq:ebt_p} defines the energy at the level of a single generation step. A complete output of the agent is a token sequence $y=(y_1,\ldots,y_T)$ from which a semantic state $g=\mathrm{state}(y)$ is extracted. Extending the Boltzmann prescription to the sequence level, we posit
\begin{align}
  p_\theta(y\,|\,x) \;\propto\; e^{-E_\theta(x,\,y)},
  \label{eq:ebt_seq}
\end{align}
where $E_\theta(x,y)$ is a sequence-level energy and $Z_\theta(x):=\sum_{y}e^{-E_\theta(x,y)}$ is the corresponding partition function. For an autoregressive EBT the precise relationship between this sequence-level energy and the per-step energy in Eq.~\eqref{eq:ebt_p} involves the context-dependent normalization at each decoding step and is model-specific; the key point for the coarse-graining argument below is only the Boltzmann form of Eq.~\eqref{eq:ebt_seq}. The state-level transition kernel is then obtained by summing over all token sequences that realize $g$:
\begin{align}
  \mathcal{T}(g\!\leftarrow\!f) = \sum_{y:\,\mathrm{state}(y)=g} p_\theta(y\,|\,x_f)
  = \frac{Z_\theta(x_f,\,g)}{Z_\theta(x_f)},
  \label{eq:ebt_T}
\end{align}
where $Z_\theta(x_f,g):=\sum_{y:\,\mathrm{state}(y)=g}e^{-E_\theta(x_f,y)}$ is a partial partition function restricted to the microstate sector of state $g$. Defining the conditional free energy
\begin{align}
  F(g\,|\,f) \;:=\; -\log Z_\theta(x_f,\,g),
\end{align}
which absorbs both the energetic and entropic contributions from all token sequences that realize state $g$ given prompt $f$, Eq.~\eqref{eq:ebt_T} becomes $\mathcal{T}(g\!\leftarrow\!f) = e^{-F(g|f)}/Z_\theta(x_f)$. Substituting into the detailed balance condition Eq.~(5) of the main text, $\log[\mathcal{T}(g\!\leftarrow\!f)/\mathcal{T}(f\!\leftarrow\!g)] = \beta V(f) - \beta V(g)$, gives
\begin{align}
  F(f\,|\,g) - F(g\,|\,f)
  = \beta\bigl[V(f)-V(g)\bigr]
  + \bigl[\log Z_\theta(x_f) - \log Z_\theta(x_g)\bigr].
  \label{eq:ebt_constraint}
\end{align}
Since $\beta V$ and $\log Z_\theta$ are both state functions, the right-hand side is an exact 1-form: a difference $h(g)-h(f)$ with $h(f):=-\beta V(f)-\log Z_\theta(x_f)$. This reveals the physical meaning of $V(f)$: it is a \emph{macroscopic free energy} at the level of semantic states, obtained by coarse-graining the token-level energy $E_\theta$ over all microstates that realize state $f$. In this identification $\beta$ inherits a concrete meaning: it is the proportionality constant that converts between the dimensionless token-level energy $E_\theta$ in Eq.~\eqref{eq:ebt_p} and the state-level potential $V$. Detailed balance constrains only the \emph{antisymmetric} part of the conditional free energy: writing $F(g\,|\,f) = S(f,g) + A(f,g)$ with $S$ symmetric and $A$ antisymmetric, the requirement is
\begin{align}
  A(f,g) \;=\; \tfrac{1}{2}\bigl[h(f)-h(g)\bigr],
  \label{eq:ebt_exact_form}
\end{align}
i.e.\ the antisymmetric part must be an exact 1-form on the state graph. The symmetric part $S(f,g)$, which encodes the mutual difficulty of translating between states $f$ and $g$ regardless of direction, is completely unconstrained by detailed balance.

\begin{figure}[tbp]
    \centering
    \includegraphics[width=0.9\linewidth]{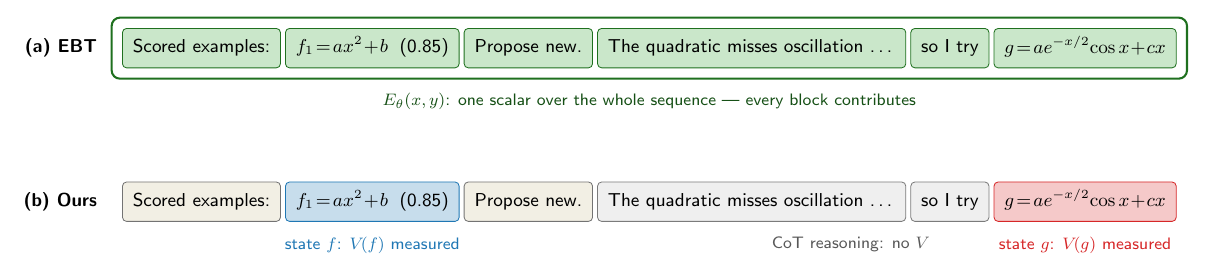}
    \caption{Resolution contrast between the microscopic architectural energy and the macroscopic state-level potential. (a)~In an energy-based model such as EBT, $E_\theta(x,\hat y)$ is a per-token energy; inference proceeds by gradient-descent energy minimization, and the full sequence is still generated autoregressively. (b)~Our potential $V(f)$ is a macroscopic free energy defined on the semantically closed states $f$ (blue) and $g$ (red), obtained by coarse-graining over all token sequences that realize a given state. The intermediate chain-of-thought reasoning tokens (gray) carry no $V$.}
    \label{fig:ebt_vs_ours}
\end{figure}

Crucially, the EBM framework alone does not guarantee Eq.~\eqref{eq:ebt_exact_form}. In an EBT, for instance, the input $x$ and output $\hat y$ are structurally asymmetric (different positional encodings, attention masks, and contexts), so the antisymmetric part of $F$ is generically nonzero and need not factorize into a difference of state functions. Whether a given $E_\theta$ induces a global potential $V$ consistent with detailed balance is therefore an empirical question, and applying our measurement framework to energy-based models is not essentially different from applying it to autoregressive models. Figure~\ref{fig:ebt_vs_ours} illustrates the difference in resolution between the two levels of description.

\end{document}